\documentclass[3p,times]{elsarticle}
\usepackage{amsmath}
\usepackage{xcolor}
\usepackage{booktabs}
\usepackage{subcaption}
\usepackage{verbatim}
\usepackage{multirow}
\usepackage{lineno}
\usepackage{url}
\usepackage{array}

\journal{arXiv}
\begin{document}

\begin{frontmatter}
\title {Two-stage Progressive Residual Dense Attention Network for Image Denoising}

\author[mymainaddress]{Wencong Wu}

\author[mymainaddress]{An Ge}

\author[mymainaddress]{Guannan Lv}

\author[mymainaddress]{Yuelong Xia}

\author[mymainaddress]{Yungang Zhang\corref{mycorrespondingauthor}}
\ead{yungang.zhang@ynnu.edu.cn}

\author[mymainaddress]{Wen Xiong\corref{mycorrespondingauthor}}
\ead{wen.xiong@ynnu.edu.cn}

\cortext[mycorrespondingauthor]{Corresponding author}

\address[mymainaddress]{School of Information Science and Technology, Yunnan Normal University, Kunming 650500, Yunnan Province, China}

\begin{abstract}
Deep convolutional neural networks (CNNs) for image denoising can effectively exploit rich hierarchical features and have achieved great success. However, many deep CNN-based denoising models equally utilize the hierarchical features of noisy images without paying attention to the more important and useful features, leading to relatively low performance. To address the issue, we design a new Two-stage Progressive Residual Dense Attention Network (TSP-RDANet) for image denoising, which divides the whole process of denoising into two sub-tasks to remove noise progressively. Two different attention mechanism-based denoising networks are designed for the two sequential sub-tasks: the residual dense attention module (RDAM) is designed for the first stage, and the hybrid dilated residual dense attention module (HDRDAM) is proposed for the second stage. The proposed attention modules are able to learn appropriate local features through dense connection between different convolutional layers, and the irrelevant features can also be suppressed. The two sub-networks are then connected by a long skip connection to retain the shallow feature to enhance the denoising performance. The experiments on seven benchmark datasets have verified that compared with many state-of-the-art methods, the proposed TSP-RDANet can obtain favorable results both on synthetic and real noisy image denoising. The code of our TSP-RDANet is available at https://github.com/WenCongWu/TSP-RDANet.
\end{abstract}


\end{frontmatter}

\section{Introduction}
Image denoising is a hot topic in computer vision tasks, which is a key preparatory work for the subsequent high-level tasks so that the performance of these tasks can be improved. Image denoising seeks to solve an inverse problem to obtain the `clear' image from an image contaminated by noise, and it can be described as $x = y - N$ for the additive white Gaussian noise (AWGN), where $x$, $y$, and $N$ denote the denoised `clear' image, the noisy image, and the noise, respectively. Many image denoising methods have been designed in recent decades. For example, Dabov et al. \cite{Dabov2007} presented the block-matching and 3-D filtering (BM3D) for image denoising, which applies the non-local self-similarity (NSS) and an enhanced sparse learning scheme in the transform domain to enhance model performance. The trilateral weighted sparse coding (TWSC) \cite{Xu2018} was proposed for practical denoising scenes. Later, Ou et al. \cite{Ou2022} designed a weighted group sparse coding model using multi-scale NSS to improve the denoising performance. Some researchers have also tried to tackle the real-world noise removal task. For example, the multi-channel weighted nuclear norm minimization (MCWNNM) \cite{Xu2017} was developed for noise removal in real scenes. Although these methods can obtain impressive denoising performance, they need to manually set parameters for different denoising tasks, and they are generally equipped with complex optimization algorithms, which result in high computational cost.

Various deep neural networks (DNNs) based denoising models have been proposed in recent years. Compared with the above mentioned traditional denoising techniques, the DNN-based denoising models need fewer model hyper-parameters, and generally have faster denoising speeds. For example, Zhang et al. \cite{Zhang2017} presented the denoising CNN (DnCNN), which surpasses most of the traditional methods. Peng et al. \cite{Peng2019} designed the dilated residual network, named as the DSNet, where the symmetric skip connection is applied in different convolutional layers to extract hierarchical features, the denoising results of the DSNet therefore are improved. Zhang et al. \cite{Zhang2021} developed the residual dense network (RDN) to restore the degraded image, which uses the densely connected layer to capture rich hierarchical features. Jia et al. \cite{Jia2021} proposed the dense dense U-Net (DDUNet) for removing the noise from noisy images, which adopts the multi-scale dense connection to obtain more image features. Although these DNN-based models are able to extract rich features to achieve favorable denoising performance, they treat all the extracted features equally, which may be an obstacle to improving model performance.

To better utilize the more useful and informative image features, several attention-based denoising methods have been developed. Tian et al. \cite{TianX2020} presented an attention-guided denoising CNN (ADNet) to eliminate the Gaussian and real noise, and the model contains an attention block to capture the noise information in the noisy images with complex backgrounds. Anwar et al. \cite{Anwar2019} designed a real image denoising network (RIDNet), which exploits the feature attention block to capture the channel dependencies to further improve denoising quality.

Recently, progressive models have also been developed to promote image denoising effect. Zamir et al. \cite{Zamir2020} presented a multi-stage architecture MPRNet for image restoration, which utilizes encoder-decoders to capture multi-scale features, and the supervised attention module is developed to refine the filtered features and the degraded images in each stage. Bai et al. \cite{Bai2023} developed a progressive denoising network (MSPNet), which decomposes the overall denoising process into multiple sub-steps to improve noise reduction gradually. Although the progressive denoising networks such as the MPRNet and MSPNet can improve denoising performance, the complex network structures bring large numbers of parameters, and their application in real denoising scenarios is therefore limited.

It also can be seen that in the current image restoration area, the Transformers-based models can generally outperform most CNN-based models, nevertheless some researchers have pointed out that the CNN-based models can also reach or even surpass transformers as appropriate network structures and components can be provided \cite{Wang2023}. Moreover, the transformer-based models generally suffer from the problems of large model scales and difficulty in training. Instead, the CNN-based models are relatively easier to be deployed on various terminal devices with limited computing resources. Therefore, the CNN-based denoising models still enrich great research potential.

Motivated by the promising performance of the progressive denoising scheme and the attention-based feature learning, we propose a two-stage progressive residual dense attention network for image denoising, named as the TSP-RDANet, which uses two heterogeneous networks to enhance feature interaction and feature representation ability. The residual dense attention module (RDAM) and the hybrid dilated residual dense attention module (HDRDAM) are developed to capture rich hierarchical features, respectively. The residual learning is embedded into these modules to promote denoising performance, and the network training can be accelerated as well. The long skip connections are employed between these attention modules to infuse the relevant and useful features. The major contributions of this work are listed as follows:

(1) We novelly design the residual dense attention module (RDAM) and the hybrid dilated residual dense attention module (HDRDAM). These two different attention modules are utilized in our two-stage progressive denoising models. The RDAM and HDRDAM use densely connected layers to extract rich local features, in which the irrelevant features are filtered by the attention mechanism, and the residual learning is applied both in the RDAM and HDRDAM to enhance the denoising performance of the network.

(2) A novel two-stage progressive residual dense attention network (TSP-RDANet) is proposed for image denoising, which decomposes the entire denoising process into two sub-tasks to progressively restore a noisy image.

(3) Experiments on multiple synthetic and real-world datasets have verified that our TSP-RDANet obtains promising denoising performance compared with many other state-of-the-art models.

The rest of this paper has the following organization. Section \ref{Related_work} shows the related techniques for image denoising. In Section \ref{Proposed_method}, we introduce the proposed TSP-RDANet model and the proposed attention modules. Section \ref{Experiment} offers our experimental details and results. The conclusion is reported in Section \ref{Conclusion}.

\section{Related work}\label{Related_work}

\subsection{CNN based image denoising techniques}
During the past years, a number of image denoising methods based on convolutional neural networks have been developed \cite{Zhang2017}. For instance, a denoising CNN (DnCNN) model was proposed by Zhang et al. \cite{Zhang2017}, which uses rectified linear units (ReLU) \cite{Krizhevsky2012}, batch normalization (BN) \cite{Ioffe2015}, and residual learning \cite{He2016} to speed up its training procedure and promote model performance. The image restoration CNN (IRCNN) \cite{ZhangZGZ2017} utilizes the dilated convolution \cite{Yu2015} to enlarge its receptive field to obtain more contextual information. Peng et al. \cite{Peng2019} applied the symmetric skip connection and dilated convolution to develop the dilated residual networks (DSNet) for image denoising. The FFDNet \cite{Zhang2018} introduces an adjustable noise level map to increase the flexibility of Gaussian noise removal, and the FFDNet employs the downsampled sub-images to achieve fast image denoising.

Generally, the image denoising methods based on CNN can obtain better performance than many traditional denoising methods. However, the early CNN-based denoising models are mainly developed for synthetic noise removal, and it is inconvenient to use them in practical denoising scenes. Therefore, many blind denoising methods have been proposed, including the VDN \cite{Yue2019}, AINDNet \cite{Kim2020}, BUIFD \cite{Helou2020}, CBDNet \cite{Guo2019}, VDIR \cite{Soh2022}, DCBDNet \cite{WuS2023}. Many of these blind denoising models contain a noise estimation sub-network to estimate noise distribution to achieve flexibility in practical image denoising, but the noise estimation networks increase the complexity of the models. Other blind denoising models without the noise estimation sub-networks have also been developed. For instance, the SADNet \cite{Chang2020} is a spatial-adaptive model, where the deformable convolution \cite{Dai2017, Zhu2019} is applied to promote its performance. Quan et al. \cite{Quan2021} presented a complex-valued denoising network (CDNet), which investigates the advantages of complex-valued operations in CNN. Li et al. \cite{Li2022} developed the AirNet, which adopts a single AirNet model to achieve the recovery of multiple types of degraded images, and excellent restoration performance was produced.

Some researchers have tried to expand the width of CNN denoising models instead of increasing network depth. For instance, Pan et al. \cite{Pan2022} presented a dual CNN (DualCNN) for low-level vision tasks, which uses two different branches to restore the main parts and details of the degraded images, respectively. The restored structures and details from the two branches are then fused to predict the recovery images. A batch-renormalization \cite{Ioffe2017} denoising network (BRDNet) was presented by Tian et al. \cite{Tian2020}, where two sub-networks were combined to extract the complementary features. Later, a dual denoising network (DudeNet) \cite{Tian2021} was developed by the same author. A dual adversarial network (DANet) was designed in \cite{Yue2020} for practical noise removal. It can be found that the dual network structure can be an effective way to improve denoising performance.

\subsection{Attention}
Different attention mechanisms are also widely used in image denoising models, which can obtain more effective features and further promote the denoising performance. The RIDNet proposed by Anwar et al. \cite{Anwar2019} employs the feature attention to capture the channel relationships in order to suppress irrelevant features. The ADNet \cite{TianX2020} utilizes attention-guided feature learning to improve its denoising result. Jiang et al. \cite{Jiang2023} developed an enhanced frequency fusion network (EFF-Net) for noise removal, in which a dynamic hash attention was designed to enhance its denoising performance. Ren et al. \cite{Ren2021} presented the DeamNet, where the dual element-wise attention mechanism module is designed to promote the denoising effect of the network. Zamir et al. \cite{Zamir2020} presented the CycleISP framework for practical image denoising, which employs spatial attention \cite{Woo2018} and channel attention \cite{Hu2018} mechanisms to exploit the inter-channel and inter-spatial dependencies. Wu et al. \cite{WuLZ2023} proposed a dual residual attention network for obtaining the denoised image, where the dual-branch structure was used to capture complementary features, and the attention mechanism in \cite{Hu2018, Woo2018} was adopted to filter unimportant features. Huang et al. \cite{Huang2022} designed a prior-guided dynamic tunable network to achieve real-world noise removal, where the global spatial and channel attention was proposed to capture non-local features. Zhuge et al. \cite{Zhuge2023} presented an enhanced feature denoising network, where the cross-channel attention was embedded into the network to enhance the interaction of channel features. Thakur et al. \cite{Thakur2023} designed a blind Gaussian denoising network, which applied a multi-scale pixel attention to capture salient features from multiple scales.

\subsection{Progressive denoising models}
Much previous work has demonstrated that compared with the single-stage counterparts, the multi-stage/progressive models can achieve more effective performance in image restoration, such as image denoising \cite{Zamir2020, Bai2023, Tian2023}, image deraining \cite{Zheng2019, Ren2019}, and image deblurring \cite{Nah2017, ZhangD2019, Suin2020}. For example, A multi-stage progressive image restoration framework (MPRNet) was developed in \cite{Zamir2020}, which adopts the supervised attention module between different stages to achieve progressive learning. Bai et al. \cite{Bai2023} designed a multi-stage progressive denoising network (MSPNet), which divides the overall denoising process into three sub-steps, and excellent noise reduction results were reported. Tian et al. \cite{Tian2023} developed a multi-stage CNN with the wavelet transform (MWDCNN), which utilizes the enhanced residual dense architectures to capture enough features for image denoising. A progressive recurrent network (PReNet) was presented in \cite{Ren2019} for image deraining, where a recurrent layer was applied to capture the feature dependencies between different stages. Fu et al. \cite{Fu2020} developed a lightweight pyramid network (LPNet) for rain removal, where the laplacian pyramid was used to generate the derained images through multiple different sub-networks. Nah et al. \cite{Nah2017} developed a multi-scale progressive network architecture for image deblurring, where the Gaussian pyramid was utilized to generate downsampled blurry images, and these images were fed into progressive sub-networks to produce the deblurred images. Zhang et al. \cite{ZhangD2019} designed a deep multi-patch hierarchical network for blur removal, which decomposes the blurred image into multiple patches gradually, and the image patches are sent into different encoder-decoders to generate the deblurred result.

\section{Proposed method}\label{Proposed_method}
In this section, the architecture of our progressive denoising model TSP-RDANet is illustrated. The proposed residual dense attention module (RDAM) and the hybrid dilated residual dense attention module (HDRDAM) are also introduced in details.

\subsection{Architecture of the TSP-RDANet}
\begin{figure*}[htbp]
	\begin{center}
		\includegraphics[width=\textwidth]{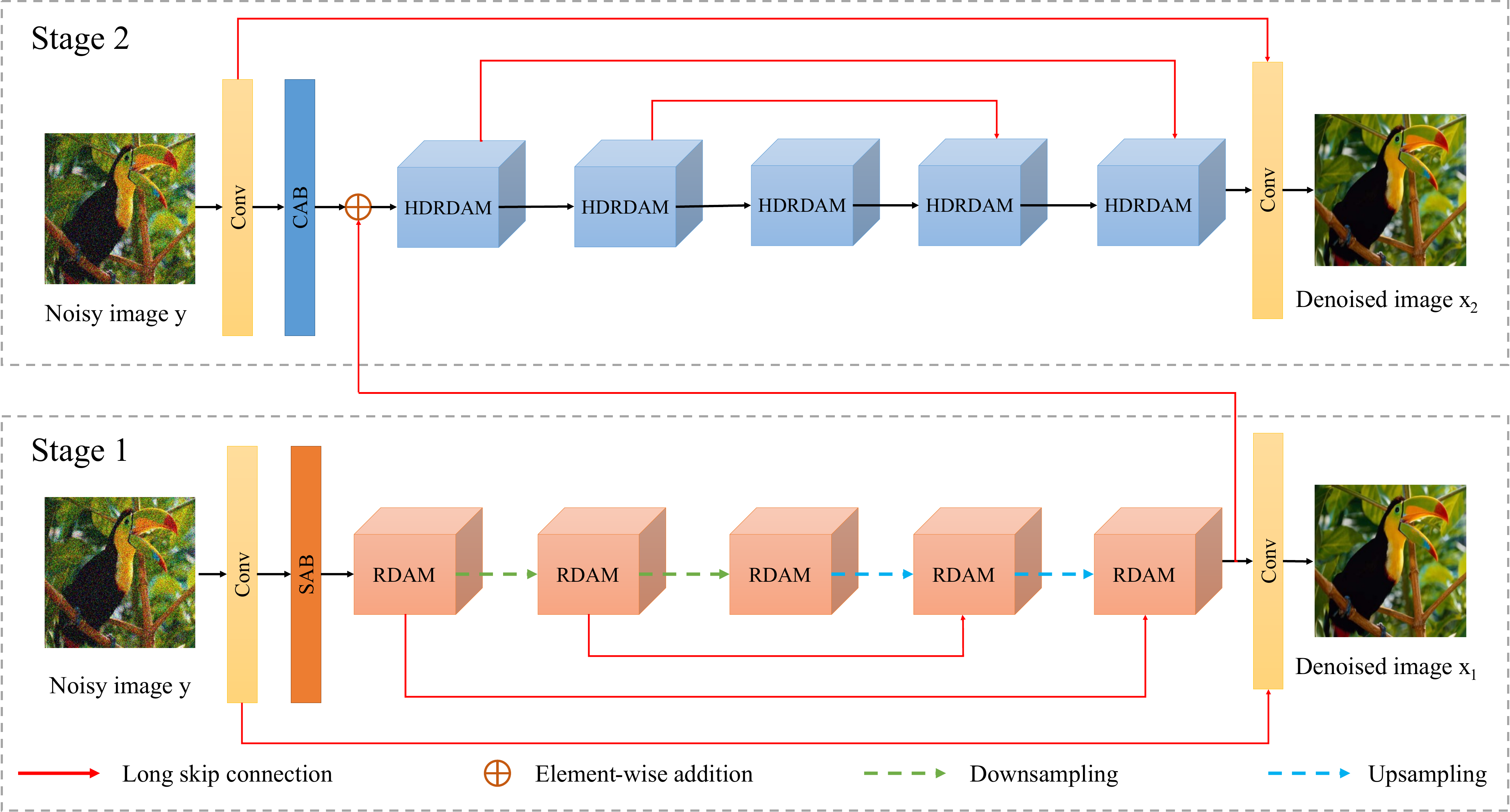}
		\caption{The architecture of the proposed TSP-RDANet.}
		\label{fig:TSP-RDANet}
	\end{center}
\end{figure*}

The framework of our TSP-RDANet is displayed in Fig. \ref{fig:TSP-RDANet}, which mainly comprises two stages: Stage 1 and Stage 2. The TSP-RDANet can extract rich image features through densely connected modules, including RDAMs and HDRDAMs. To obtain a desirable trade-off between the network performance and complexity, five RDAMs and five HDRDAMs are used for the first and the second stages denoising network, respectively. The strided convolutions and the transpose convolutions between the RDAMs are used for obtaining the multi-scale information and enlarging the receptive field. In HDRDAMs, the dilated convolutions are utilized to expand the receptive field to obtain more contextual information as well. In addition, the TSP-RDANet model adopts long skip connections to fuse the shallow and deep convolutional features, allowing the model to fully utilize the features, thereby improving its repair performance. Furthermore, the long skip connection enables the sub-networks of the two stages to interact on the different salient features to promote the expressive ability of the model.

The whole denoising process of the TSP-RDANet model is represented as follows:

\begin{equation}
    x_1, x_2 = \mathcal{F}{_{TSP-RDANet}}(y),
\end{equation}

\noindent where $x_1$ and $x_2$ denote the denoised image from different stages, and $y$ represents the noisy image. Specifically, the first stage denoising network contains two convolutional layers, a spatial attention block (SAB), five RDAMs, long skip connections, the $2 \times 2$ strided convolution (SConv), and $2 \times 2$ transposed convolution (TConv). The SConv and TConv are respectively applied for image downsampling and upsampling to augment the receptive field of the denoising network, and the multi-scale features can be extracted as well. The denoising procedure of the first stage can be formulated as follows:

\begin{equation}
    \begin{aligned}
        O_{Conv}^1 &= K * y, \\
        O_{SAB} &= SAB(O_{Conv}^1), \\
        O_{RDAM}^1 &= RDAM(O_{SAB}), \\
        O_{RDAM}^2 &= RDAM(SConv(O_{RDAM}^1)), \\
        O_{RDAM}^3 &= RDAM(SConv(O_{RDAM}^2)), \\
        O_{RDAM}^4 &= RDAM(TConv(O_{RDAM}^3) + O_{RDAM}^2), \\
        O_{RDAM}^5 &= RDAM(TConv(O_{RDAM}^4) + O_{RDAM}^1), \\
        x_1 &= K * (O_{RDAM}^5 + O_{Conv}^1),
    \end{aligned}
\end{equation}

\noindent where $O_{Conv}^1$ and $*$ denote the output of a convolutional layer and convolution operation respectively, and $K$ is the standard convolutional kernel to expand the number of the feature maps. Moreover, $O_{SAB}$ and $O_{RDAM}^j$ ($j \in$ \{1, 2, 3, 4, 5\}) is the output of the SAB and the RDAM, respectively.

The second stage denoising network comprises two convolutional layers, a channel attention block (CAB), five HDRDAMs, and long skip connections. The denoising procedure of Stage 2 can be defined as follows:

\begin{equation}
    \begin{aligned}
        O_{Conv}^2 &= K * y, \\
        O_{CAB} &= CAB(O_{Conv}^2), \\
        O_{HDRDAM}^1 &= HDRDAM(O_{CAB} + O_{RDAM}^5), \\
        O_{HDRDAM}^2 &= HDRDAM(O_{HDRDAM}^1), \\
        O_{HDRDAM}^3 &= HDRDAM(O_{HDRDAM}^2), \\
        O_{HDRDAM}^4 &= HDRDAM(O_{HDRDAM}^3 + O_{HDRDAM}^2), \\
        O_{HDRDAM}^5 &= HDRDAM(O_{HDRDAM}^4 + O_{HDRDAM}^1), \\
        x_2 &= K * (O_{HDRDAM}^5 + O_{Conv}^2),
    \end{aligned}
\end{equation}

\noindent where $O_{CAB}$ and $O_{HDRDAM}^i$ ($i \in$ \{1, 2, 3, 4, 5\}) are the output of the CAB and the HDRDAM respectively, and $O_{Conv}^2$ denotes the output of a convolutional layer. The different hierarchical features are fused by the long skip connections between the proposed attention modules, and the restoration effect therefore can be improved.

\subsection{The residual dense attention module}
\begin{figure*}[htbp]
	\begin{center}
		\includegraphics[width=\textwidth]{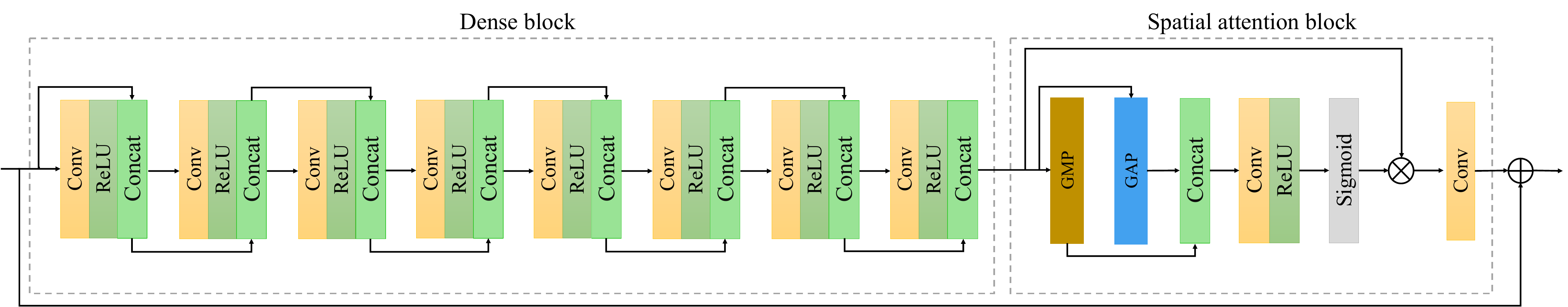}
		\caption{The structure of the designed RDAM.}
		\label{fig:RDAM}
	\end{center}
\end{figure*}

The structure of the residual dense attention module (RDAM) is shown in Fig. \ref{fig:RDAM}, which is applied for the first stage of the proposed TSP-RDANet. The RDAM contains the dense block (DB), spatial attention block (SAB) \cite{Woo2018}, and residual learning (RL) \cite{He2016}. The feature maps $f_{RDAM}$ processed by the RDAM can be expressed as follows:

\begin{equation}
    \begin{aligned}
        O_{RDAM} &= RDAM(f_{RDAM}), \\
                 &= SAB(DB(f_{RDAM})) + f_{RDAM},
    \end{aligned}
\end{equation}

\noindent where $O_{RDAM}$ is the output of the RDAM.

The DB (dense block) consists of eight standard convolutions (Conv), eight rectified linear units (ReLU) \cite{Krizhevsky2012}, and eight concatenation operations (Concat). Specifically, these convolutional layer extracts rich local and hierarchical features, and these features are non-linearly transformed by using the ReLU, then fused by the Concat. Since all the hierarchical features extracted by the DB are treated equally, the important image features are not paid extra attention, which may result in the degradation of the image restoration performance.

The spatial attention block (SAB) is therefore used here to solve the issue, which can disclose the inter-relationship of spatial features. More importantly, the SAB can give more attention on the essential image features and filter the irrelevant ones. The SAB is composed of the Conv, `$\otimes$', Concat, GAP (Global Average Pooling), ReLU, GMP (Global Max Pooling), and Sigmoid \cite{Han1995} activation. The symbols `$\otimes$' and `$\oplus$' in Fig. \ref{fig:RDAM} denote the element-wise product and the RL, respectively.

\subsection{The hybrid dilated residual dense attention module}
\begin{figure*}[htbp]
	\begin{center}
		\includegraphics[width=\textwidth]{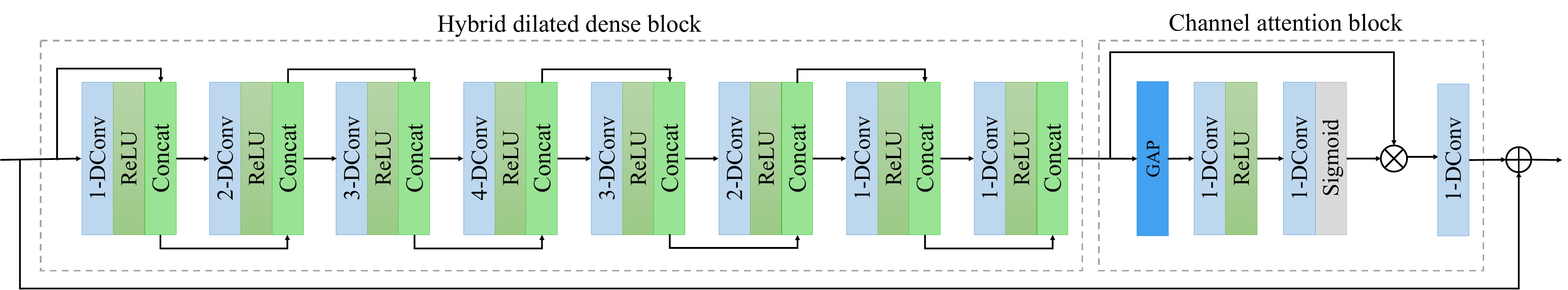}
		\caption{The structure of the designed HDRDAM.}
		\label{fig:HDRDAM}
	\end{center}
\end{figure*}

The structure of the designed hybrid dilated residual dense attention module (HDRDAM) is presented in Fig. \ref{fig:HDRDAM}, which is used in the Stage 2 of our denoising network. The HDRDAM includes the hybrid dilated dense block (HDDB), channel attention block (CAB) \cite{Hu2018}, and residual learning (RL) \cite{He2016}. The process of the feature maps $f_{HDRDAM}$ passing through the HDRDAM can be formulated as follows:

\begin{equation}
    \begin{aligned}
        O_{HDRDAM} &= HDRDAM(f_{HDRDAM}), \\
                 &= CAB(HDDB(f_{HDRDAM})) + f_{HDRDAM},
    \end{aligned}
\end{equation}

\noindent where $O_{HDRDAM}$ is the output of the HDRDAM.

The HDDB contains eight hybrid dilated convolutions ($r$-DConv in Fig. \ref{fig:HDRDAM}) \cite{Yu2015}, eight rectified linear units (ReLU) \cite{Krizhevsky2012}, and eight concatenation operations (Concat), where the `$r$' denotes the dilated rate and the range of its value is [1, 4]. Like the RDAM, the HDRDAM is also able to extract rich local features via dense connection between the dilated convolutional layers. Moreover, the hybrid dilated convolution can not only capture more useful information by enlarging the receptive field, but also can effectively remove the possible gridding phenomenon \cite{Wang2018}.

The CAB contains the GAP, 1-DConv, ReLU, Sigmoid and `$\otimes$', where the 1-DConv and the symbol `$\otimes$' in Fig. \ref{fig:HDRDAM} represent the dilated convolution with the dilated rate 1 and the element-wise product, respectively. The CAB is utilized for capturing the inter-dependencies between the spatial features, and it is also used for suppressing the unimportant features. The symbol `$\oplus$' in Fig. \ref{fig:HDRDAM} denotes the RL.

\subsection{Loss function}
We utilize different loss functions for synthetic noise removal and real noise elimination. For the Gaussian denoising, we use the loss function $L_{mse}$ that calculates the difference between the recovery image $x_i$ ($i \in$ \{1, 2\}) and the ground-truth image $x_{gt}$, which can be denoted as:

\begin{equation}
\mathcal{L} = \sum_{i=1}^2 L_{mse}(x_i, x_{gt}),
\end{equation}

\noindent where $L_{mse}$ denotes the mean squared error (MSE). In each denoising stage, the $L_{MSE}$ can be designed as:

\begin{equation}
\mathcal{L} = \left\|x_i - x_{gt}\right\|^2, i \in \{1, 2\}
\end{equation}

For real noise removal, we refer to \cite{Zamir2021} to adopt the Charbonnier loss \cite{Lai2017} and the edge loss \cite{Jiang2020} to optimize the TSP-RDANet model, which have been verified can effectively retain image details and edge textures in the recovery image. The whole loss function is expressed as:

\begin{equation}
\mathcal{L} = \sum_{i=1}^2 [L_{char}(x_i, x_{gt}) + \lambda_{edge}L_{edge}(x_i, x_{gt})],
\label{eq:6}
\end{equation}

\noindent where $L_{Char}$ and $L_{edge}$ represent the Charbonnier loss \cite{Lai2017} and the edge loss \cite{Jiang2020}, respectively. We also set $\lambda_{edge}$ to 0.1 according to \cite{WuLZ2023}. For each stage, the Eq. (\ref{eq:6}) can be further formulated as follows:

\begin{equation}
\mathcal{L} = \sqrt{\left\|x_i - x_{gt}\right\|^2 + \epsilon^2} + 0.1 * \sqrt{\left\|\bigtriangleup{(x_i)} - \bigtriangleup{(x_{gt})}\right\|^2 + \epsilon^2}, i \in \{1, 2\}
\end{equation}

\noindent where the constant $\epsilon$ is equal to $10^{-3}$, and the $\bigtriangleup$ denotes the Laplacian operator \cite{Kamgar1999}.

\section{Experiments}\label{Experiment}
In this section, we introduce the datasets used in our experiments, the experimental settings, and the experimental results. For results comparison, the qualitative effect of the predicted images generated by different denoising models, the peak signal-to-noise ratio (PSNR), and the structural similarity index measure (SSIM) \cite{Wang2004} were utilized. In our comparative experiments, it can be noticed that the methods being compared in each task are not entirely the same, or some results of some methods are missing. This is because some compared methods did not release their code, or they could not be reproduced on our devices.

\subsection{Datasets}
For the synthetic noise reduction evaluation, the Flick2K dataset \cite{Lim2017}, which contains 2650 high-resolution color images, was adopted as the training set to train our TSP-RDANet model. To facilitate model training and accelerate the training process, these high-resolution images were randomly cut into image patches with the size of $128 \times 128$. The image patches were grayscaled to train the TSP-RDANet model for single-channel noisy removal evaluation. To produce the clean/noisy image pairs, the additive white Gaussian noise (AWGN) with different noise levels ([0, 50]) is randomly added to the ground-truth image patches. Five public benchmarks were utilized as the test datasets, including the CBSD68 \cite{Roth2005}, Kodak24 \cite{Kodak24}, McMaster \cite{Zhang2011}, Set12 \cite{Roth2005}, and BSD68 \cite{Roth2005}.

For the performance assessment of the real-world image denoising tasks, we adopted the SIDD medium dataset \cite{Abdelhamed2018} as the training dataset, which comprises 320 pairs of high-resolution noisy images and their near noise-free counterparts. These high-resolution image pairs were also randomly cropped into $128 \times 128$ image patches to facilitate the training procedure. The SIDD validation set \cite{Abdelhamed2018} and DND sRGB dataset \cite{Plotz2017} were selected as the test sets for real noise removal.

\subsection{Experimental settings}
All of our experiments were implemented on the Ubuntu 18.04 from a PC equipped with a CPU of Intel(R) Core(TM) i7-11700KF, 32GB RAM, and an Nvidia GeForce RTX 3080Ti GPU. We employed Python 3.8 and Pytorch 1.8 to code the training and testing packages of the TSP-RDANet model. It costed approximately 48 and 50 hours to train the proposed TSP-RDANet for grayscale and color images, respectively. Training the TSP-RDANet model for real noise reduction costed around 180 hours.

The Adam algorithm \cite{Kingma2014} was utilized to optimize the parameters of the TSP-RDANet. In every training batch, 4 image patches with the size of $128 \times 128$ are fed into the TSP-RDANet model for both the synthetic and real image denoising models. For the AWGN noise elimination models, the total number of iterations is $5\times10^{5}$. The learning rate ($lr$) is firstly set to $1.0\times10^{-4}$ and decreases by half after each $1\times10^{5}$ iterations. We run 120 epochs to train the real-world denoising model, and the initial $lr$ is $2.0\times10^{-4}$. Moreover, we utilized the cosine annealing strategy \cite{Loshchilov2017} to gradually reduce the $lr$ to $1.0\times10^{-6}$.

\subsection{Ablation study}
In this section, the effectiveness of different stages and different numbers of the RDAM and HDRDAM modules in the proposed TSP-RDANet model are explored. We implemented an ablation experiment on the three denoising models, including: Stage 1 only, Stage 2 only, and the whole TSP-RDANet. Moreover, we also discussed the performance impact of different numbers of RDAM and HDRDAM on the TSP-RDANet model. The average PSNR and SSIM of these denoising models on the BSD68 dataset \cite{Roth2005} are used for comparison.

\begin{table*}[htbp]
\centering
\caption{The ablation results of different stages on the BSD68 dataset.}
\label{tab:BSD68_val}
\begin{tabular}{cccc}
\hline
Models & Stage 1 only & Stage 2 only & TSP-RDANet (whole)\\
\hline
PSNR & 27.78 & 27.84 & 27.88\\
\hline
SSIM & 0.781 & 0.783 & 0.785\\
\hline
\end{tabular}
\end{table*}

In our ablation experiments, the AWGN with noise level 35 was added in the tested images, and the corresponding denoising results of three models are reported in Table \ref{tab:BSD68_val}. It can be seen that compared with Stage 1 only and Stage 2 only, the whole TSP-RDANet model achieves the best performance. In terms of the PSNR values, the denoising result of the TSP-RDANet model surpasses the other two denoising models by 0.10 dB and 0.04 dB. In terms of the SSIM values, the denoising result of the TSP-RDANet model exceeds the rest two denoising models by 0.004 and 0.002. It can be found that by utilizing a progressive strategy, the performance of the TSP-RDANet can be gradually improved.

\begin{table}[htbp]
\centering
\caption{The denoising performances (PSNR/SSIM) using different numbers of the RDAM and HDRDAM on the BSD68 dataset.}
\label{tab:num_RDAM_HDRDAM}
\begin{tabular}{ccccc}
\hline
Noise levels & TSP-RDANet$_{1\times1}$ & TSP-RDANet$_{3\times3}^1$ & TSP-RDANet$_{7\times7}^3$ & TSP-RDANet$_{5\times5}^2$ \\
\hline
$\sigma$=10 & 33.36/0.914 & 33.75/0.925 & 33.82/0.926 & 33.80/0.927\\
\hline
$\sigma$=20 & 29.91/0.840 & 30.21/0.854 & 30.29/0.857 & 30.28/0.858\\
\hline
$\sigma$=30 & 28.07/0.783 & 28.37/0.799 & 28.45/0.803 & 28.44/0.804\\
\hline
$\sigma$=40 & 26.84/0.737 & 27.15/0.756 & 27.24/0.760 & 27.23/0.761\\
\hline
$\sigma$=50 & 25.91/0.698 & 26.26/0.720 & 26.35/0.725 & 26.34/0.726\\
\hline
\end{tabular}
\end{table}

The denoising results of the TSP-RDANet model equipped with different numbers of the RDAM and HDRDAM modules were also evaluated. Table \ref{tab:num_RDAM_HDRDAM} shows the comparison of average PSNR and SSIM values at different noise levels, where $k$ in TSP-RDANet$_{k \times k}^m$ is the numbers of the RDAM and HDRDAM modules in two stages, and $m$ is the number of the downsampling and upsampling operations in Stage 1. In Table \ref{tab:num_RDAM_HDRDAM}, one can see that compared with the TSP-RDANet$_{1\times1}$ and TSP-RDANet$_{3\times3}^1$, the TSP-RDANet$_{5\times5}^2$ achieved the best denoising effects on PSNR and SSIM values. Although the TSP-RDANet$_{5\times5}^2$ is slightly lower than the TSP-RDANet$_{7\times7}^3$ on average PSNR value, its performance on average SSIM outperforms the TSP-RDANet$_{7\times7}^3$. After full consideration of model performance and complexity, we chose the TSP-RDANet$_{5\times5}^2$ as our `baseline'.

\subsection{Synthetic noise removal evaluation}\label{SNRE}
In this subsection, we give the experimental results on synthetic noisy images. The Set12 and BSD68 datasets were applied for the grayscale image denoising test. The CBSD68, Kodak24, and McMaster datasets were employed for color image noise reduction experiments. For both the grayscale and color images, the noisy images were produced by adding the AWGN with noise levels of 15, 25, and 50 respectively into the ground-truth images. In Tables \ref{tab:Set12_PSNR}-\ref{tab:BSD68}, the red and blue numbers denote the top two denoising results, respectively.

The proposed TSP-RDANet was compared with the classical and the state-of-the-art denoising models, including the BM3D \cite{Dabov2007}, IRCNN \cite{ZhangZGZ2017}, DnCNN \cite{Zhang2017}, FFDNet \cite{Zhang2018}, BUIFD \cite{Helou2020}, DudeNet \cite{Tian2021}, MWDCNN \cite{Tian2023}, ADNet \cite{TianX2020}, CDNet \cite{Quan2021}, DSNetB \cite{Peng2019}, AINDNet \cite{Kim2020}, RIDNet \cite{Anwar2019}, VDN \cite{Yue2019}, BRDNet \cite{Tian2020}, and AirNet \cite{Li2022}.

\subsubsection{Grayscale image denoising evaluation}

The PSNR values of different denoising models on the Set12 dataset are displayed in Table \ref{tab:Set12_PSNR}. One can find that compared with other models, our TSP-RDANet obtains the leading average PSNR results at all the compared noise levels, especially at the higher noise levels. The results reveal that our model is more powerful on discriminating between normal and noisy signals, which benefits from the utilization of the progressive scheme and attention-guided feature filtering.

\begin{table*}[htbp]\tiny
\centering
\caption{The PSNR comparisons of multiple denoising models on the Set12 dataset.}
\label{tab:Set12_PSNR}
\begin{tabular}{|c|c|c|c|c|c|c|c|c|c|c|c|c|c|c|}
\hline
 Noise levels & Models & C.man & House & Peppers & Starfish &  Monarch &  Airplane & Parrot &  Lena &  Barbara &  Boat &  Man & Couple & Average\\
\hline
\hline
\multirow{10}*{$\sigma$=15} & BM3D \cite{Dabov2007} & 31.91 & 34.93 & 32.69 & 31.14	& 31.85	& 31.07	& 31.37	& 34.26	& \textcolor{red}{33.10} & 32.13	& 31.92	& 31.10	& 32.37\\
\cline{2-15}
    & DnCNN-S \cite{Zhang2017} & \textcolor{blue}{32.61} & 34.97 & 33.30 & 32.20 & 33.09 & 31.70 & 31.83 & 34.62 & 32.64 & 32.42 & \textcolor{blue}{32.46} & 32.47 & 32.86\\
\cline{2-15}
    & IRCNN \cite{ZhangZGZ2017} & 32.55 & 34.89 & \textcolor{blue}{33.31}	& 32.02	& 32.82	& 31.70	& 31.84	& 34.53	& 32.43	& 32.34	& 32.40	& 32.40	& 32.77\\
\cline{2-15}
    & FFDNet \cite{Zhang2018} & 32.43	& 35.07	& 33.25	& 31.99	& 32.66	& 31.57	& 31.81	& 34.62	& 32.54	& 32.38	& 32.41	& 32.46	& 32.77\\
\cline{2-15}
    & BUIFD \cite{Helou2020} & 31.74	& 34.78	& 32.80	& 31.92	& 32.77	& 31.34 & 31.39	& 34.38	& 31.68	& 32.18	& 32.25	& 32.22	& 32.46\\
\cline{2-15}
    & DudeNet \cite{Tian2021} & \textcolor{red}{32.71} & \textcolor{blue}{35.13} & \textcolor{red}{33.38} & \textcolor{red}{32.29} & \textcolor{blue}{33.28} & \textcolor{red}{31.78} & 31.93 & \textcolor{blue}{34.66} & 32.73 & 32.46 & \textcolor{blue}{32.46} & 32.49 & \textcolor{blue}{32.94}\\
\cline{2-15}
    & MWDCNN \cite{Tian2023} & 32.53 & 35.09 & 33.29 & \textcolor{blue}{32.28} & 33.20 & \textcolor{blue}{31.74} & \textcolor{red}{31.97} & 34.64 & 32.65 & \textcolor{blue}{32.49} & \textcolor{blue}{32.46} & \textcolor{blue}{32.52} & 32.91\\
\cline{2-15}
    & TSP-RDANet & 32.55 & \textcolor{red}{35.35} & 33.28 & 32.18 & \textcolor{red}{33.30}	& 31.72 & \textcolor{blue}{31.94}	& \textcolor{red}{34.80}	& \textcolor{blue}{32.76} & \textcolor{red}{32.57}	& \textcolor{red}{32.49}	& \textcolor{red}{32.62} & \textcolor{red}{32.96} \\
\hline
\hline
\multirow{10}*{$\sigma$=25} & BM3D \cite{Dabov2007} & 29.45 & 32.85 & 30.16 & 28.56 & 29.25 & 28.42 & 28.93 & 32.07 & \textcolor{red}{30.71} & 29.90 & 29.61 & 29.71 & 29.97 \\
\cline{2-15}
    & DnCNN-S \cite{Zhang2017} & 30.18 & 33.06 & 30.87 & 29.41 & 30.28 & 29.13 & 29.43 & 32.44 & 30.00 & 30.21 & 30.10 & 30.12 & 30.43 \\
\cline{2-15}
    & IRCNN \cite{ZhangZGZ2017} & 30.08 & 33.06 & 30.88 & 29.27 & 30.09 & 29.12 & 29.47 & 32.43 & 29.92 & 30.17 & 30.04 & 30.08 & 30.38 \\
\cline{2-15}
    & FFDNet \cite{Zhang2018} & 30.10 & 33.28 & 30.93 & 29.32 & 30.08 & 29.04 & 29.44 & 32.57 & 30.01 & 30.25 & \textcolor{blue}{30.11} & \textcolor{blue}{30.20} & 30.44\\
\cline{2-15}
    & BUIFD \cite{Helou2020} & 29.42	& 33.03	& 30.48	& 29.21	& 30.20	& 28.99	& 28.94	& 32.20	& 29.18	& 29.97	& 29.88	& 29.90	& 30.12\\
\cline{2-15}
    & DudeNet \cite{Tian2021} & \textcolor{blue}{30.23} & 33.24 & \textcolor{red}{30.98} & \textcolor{blue}{29.53} & 30.44 & \textcolor{blue}{29.14} & \textcolor{blue}{29.48} & 32.52 & 30.15 & 30.24 & 30.08 & 30.15 & 30.52\\
\cline{2-15}
    & MWDCNN \cite{Tian2023} & 30.19 & \textcolor{blue}{33.33} & 30.85 & \textcolor{red}{29.66} & \textcolor{blue}{30.55} & \textcolor{red}{29.16} & \textcolor{blue}{29.48} & \textcolor{blue}{32.67} & 30.21 & \textcolor{blue}{30.28} & 30.10 & 30.13 & \textcolor{blue}{30.55}\\
\cline{2-15}
    & TSP-RDANet & \textcolor{red}{30.28} & \textcolor{red}{33.63} & \textcolor{blue}{30.95} & 29.49	& \textcolor{red}{30.58}	& \textcolor{red}{29.16} & \textcolor{red}{29.53}	& \textcolor{red}{32.80}	& \textcolor{blue}{30.36} & \textcolor{red}{30.44}	& \textcolor{red}{30.17}	& \textcolor{red}{30.37} & \textcolor{red}{30.65} \\
\hline
\hline
\multirow{10}*{$\sigma$=50} & BM3D \cite{Dabov2007} & 26.13 & 29.69 & 26.68 & 25.04 & 25.82 & 25.10 & 25.90 & 29.05 & \textcolor{red}{27.22} & 26.78 & 26.81 & 26.46 & 26.72\\
\cline{2-15}
    & DnCNN-S \cite{Zhang2017} & 27.03 & 30.00 & 27.32 & 25.70 & 26.78 & 25.87 & 26.48 & 29.39 & 26.22 & 27.20 & 27.24 & 26.90 & 27.18\\
\cline{2-15}
    & IRCNN \cite{ZhangZGZ2017} & 26.88 & 29.96 & 27.33 & 25.57 & 26.61 & 25.89 & 26.55 & 29.40 & 26.24 & 27.17 & 27.17 & 26.88 & 27.14 \\
\cline{2-15}
    & FFDNet \cite{Zhang2018} & 27.05 & 30.37 & \textcolor{blue}{27.54} & 25.75 & 26.81 & 25.89 & \textcolor{blue}{26.57} & \textcolor{blue}{29.66} & 26.45 & \textcolor{blue}{27.33} & \textcolor{blue}{27.29} & 27.08 & 27.32 \\
\cline{2-15}
    & BUIFD \cite{Helou2020} & 25.44	& 29.76	& 26.50	& 24.87	& 26.49	& 25.34	& 25.07	& 28.81	& 25.49	& 26.59	& 26.87	& 26.34	& 26.46\\
\cline{2-15}
    & DudeNet \cite{Tian2021} & \textcolor{blue}{27.22} & 30.27 & 27.51 & \textcolor{blue}{25.88} & 26.93 & 25.88 & 26.50 & 29.45 & 26.49 & 27.26 & 27.19 & 26.97 & 27.30\\
\cline{2-15}
    & MWDCNN \cite{Tian2023} & 26.99 & \textcolor{blue}{30.58} & 27.34 & 25.85 & \textcolor{blue}{27.02} & \textcolor{blue}{25.93} & 26.48 & 29.63 & 26.60 & 27.23 & 27.27 & \textcolor{blue}{27.11} & \textcolor{blue}{27.34}\\
\cline{2-15}
    & TSP-RDANet & \textcolor{red}{27.53} & \textcolor{red}{30.89} & \textcolor{red}{27.59} & \textcolor{red}{25.96} & \textcolor{red}{27.06}	& \textcolor{red}{25.97} & \textcolor{red}{26.67}	& \textcolor{red}{29.91}	& \textcolor{blue}{27.11} & \textcolor{red}{27.52}	& \textcolor{red}{27.33}	& \textcolor{red}{27.31} & \textcolor{red}{27.57} \\
\hline
\end{tabular}
\end{table*}

\begin{figure*}[htbp]
	\begin{center}
		\includegraphics[width=\textwidth]{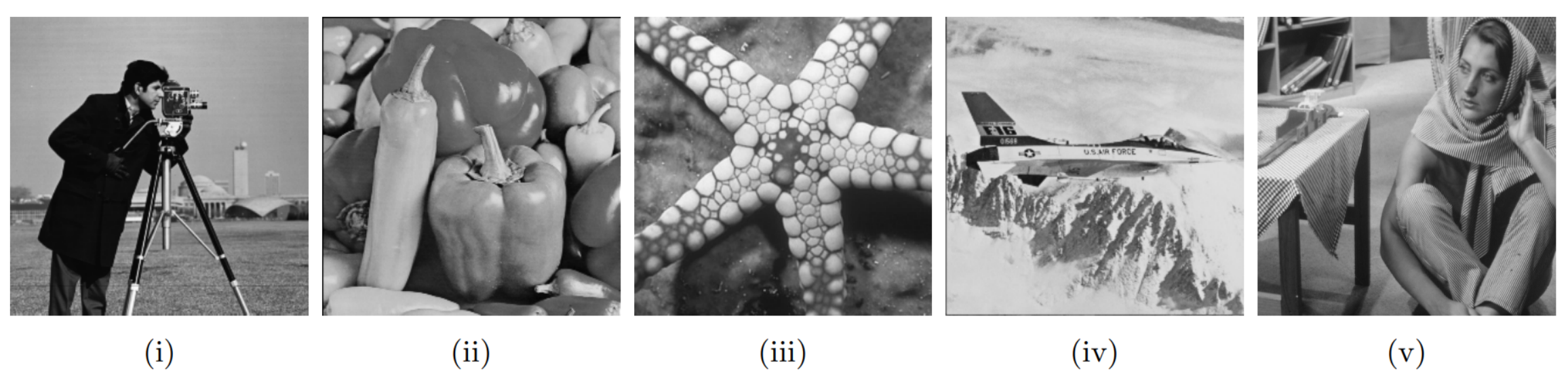}
		\caption{Five grayscale images from the Set12 dataset. (\romannumeral1) ``C.man'' image, (\romannumeral2) ``Peppers'' image, (\romannumeral3) ``Starfish'' image, (\romannumeral4) ``Airplane'' image, (\romannumeral5) ``Barbara'' image.}
        \label{fig:five_images}
	\end{center}
\end{figure*}

It is worth noting from Table \ref{tab:Set12_PSNR} that the BM3D is superior to our proposed TSP-RDANet on the ``Barbara'' image (Fig. \ref{fig:five_images} (\romannumeral5)) at noise levels 15, 25, and 50, which is we think is due to the fact that the `Barbara' image contains rich repetitive structures, that can be effectively learned by the non-local self-similarity (NSS) based method. Furthermore, the denoising effect of the DudeNet is better than that of the TSP-RDANet on the ``C.man'', ``Peppers'', ``Starfish'', and ``Airplane'' images ( Fig. \ref{fig:five_images} (\romannumeral1)-(\romannumeral4)) at noise level 15. It can be found from Fig. \ref{fig:five_images} (\romannumeral1)-(\romannumeral4) that these images contain many low-frequency regions, and their texture structures are not severely damaged under weak noise, which are beneficial to the DudeNet using the dual-branch structure to predict the noise map and obtain high-quality denoised image. Additionally, the proposed TSP-RDANet model has a larger receptive field, which may make the model insensitive to the weak noise in the low-frequency image areas.

Table \ref{tab:Set12_SSIM} shows the averaged SSIM results of the compared methods on the Set12 dataset, and the noisy images at the noise levels of 15, 25, and 50 were used for evaluation. The proposed TSP-RDANet shares the leading position with the ADNet at the noise level 15. It can also be discovered that our model surpasses other compared methods at the more challenging higher noise levels, which again verifies the discrimination ability of our model.

\begin{table}[htbp]\footnotesize
\centering
\caption{The SSIM comparisons of multiple denoising models on the Set12 dataset.}
\label{tab:Set12_SSIM}
\begin{tabular}{ccccccccc}
\hline
Models & BM3D \cite{Dabov2007} & IRCNN  \cite{ZhangZGZ2017} & DnCNN-S \cite{Zhang2017} & FFDNet \cite{Zhang2018} & BUIFD \cite{Helou2020} & ADNet \cite{TianX2020} & CDNet \cite{Quan2021} & TSP-RDANet \\
\hline
$\sigma$=15 & 0.896  & 0.901 & \textcolor{blue}{0.903} & \textcolor{blue}{0.903} & 0.899 & \textcolor{red}{0.905} & \textcolor{blue}{0.903} & \textcolor{red}{0.905} \\
\hline
$\sigma$=25 & 0.851 & 0.860  & 0.862 & 0.864 & 0.855 & \textcolor{blue}{0.865} & \textcolor{blue}{0.865} & \textcolor{red}{0.867} \\
\hline
$\sigma$=50 & 0.766 & 0.780 & 0.783  & 0.791 & 0.755 & 0.791 & \textcolor{blue}{0.792} & \textcolor{red}{0.798} \\
\hline
\end{tabular}
\end{table}

The BSD68 dataset \cite{Roth2005} was also used to evaluate grayscale image denoising performance at noise levels 15, 25, and 50. Table \ref{tab:BSD68} reports the averaged PSNR and SSIM values of the compared denoising methods. It can be seen that the proposed TSP-RDANet achieves the best performance at noise levels 25 and 50 than other denoising methods, and obtains competitive results at noise level 15.

\begin{table}[htbp]
\centering
\caption{The PSNR and SSIM comparisons of image denoising models on the BSD68 dataset.}
\label{tab:BSD68}
\begin{tabular}{ccccc}
\hline
Metrics & Models & $\sigma$=15 & $\sigma$=25 & $\sigma$=50\\
\cline{1-5}
\multirow{14}*{PSNR} & BM3D \cite{Dabov2007} & 31.07 & 28.57 & 25.62\\
\cline{2-5}
    & DnCNN-S \cite{Zhang2017} & 31.72 & 29.23 & 26.23\\
\cline{2-5}
    & IRCNN \cite{ZhangZGZ2017} & 31.63 & 29.15 & 26.19 \\
\cline{2-5}
    & FFDNet \cite{Zhang2018} & 31.63	& 29.19 & 26.29\\
\cline{2-5}
& BUIFD \cite{Helou2020} &  31.35 & 28.75 & 25.11\\
\cline{2-5}
    & DSNetB \cite{Peng2019} & 31.69 & 29.22 & 26.29 \\
\cline{2-5}
    & ADNet \cite{TianX2020} & 31.74 & 29.25 & 26.29 \\
\cline{2-5}
    & AINDNet \cite{Kim2020} & 31.69 & 29.26 & 26.32 \\
\cline{2-5}
    & CDNet  \cite{Quan2021} & 31.74 & 29.28 & \textcolor{blue}{26.36} \\
\cline{2-5}
    & DudeNet \cite{Tian2021} & \textcolor{red}{31.78} & \textcolor{blue}{29.29} & 26.31 \\
\cline{2-5}
    & MWDCNN \cite{Tian2023} & \textcolor{blue}{31.77} & 29.28 & 26.29 \\
\cline{2-5}
    & TSP-RDANet & \textcolor{blue}{31.77} & \textcolor{red}{29.34} & \textcolor{red}{26.45}\\
\hline
\multirow{11}*{SSIM} & BM3D \cite{Dabov2007} & 0.872 & 0.802 & 0.687\\
\cline{2-5}
    & DnCNN-S \cite{Zhang2017} & \textcolor{blue}{0.891} & 0.828 & 0.719 \\
\cline{2-5}
    & IRCNN  \cite{ZhangZGZ2017} & 0.888 & 0.825 & 0.717 \\
\cline{2-5}
    & FFDNet \cite{Zhang2018} & 0.890 & 0.830 & 0.726 \\
\cline{2-5}
    & BUIFD \cite{Helou2020} &  0.886 & 0.819 & 0.682 \\
\cline{2-5}
    & ADNet  \cite{TianX2020} & \textcolor{red}{0.892} & 0.829 &  0.722 \\
\cline{2-5}
    & CDNet \cite{Quan2021} & \textcolor{red}{0.892} & \textcolor{blue}{0.831} & \textcolor{blue}{0.727} \\
\cline{2-5}
    & TSP-RDANet & \textcolor{red}{0.892} & \textcolor{red}{0.832} & \textcolor{red}{0.731} \\
\hline
\end{tabular}
\end{table}

\begin{figure*}[htbp]
	\begin{center}
		\includegraphics[width=\textwidth]{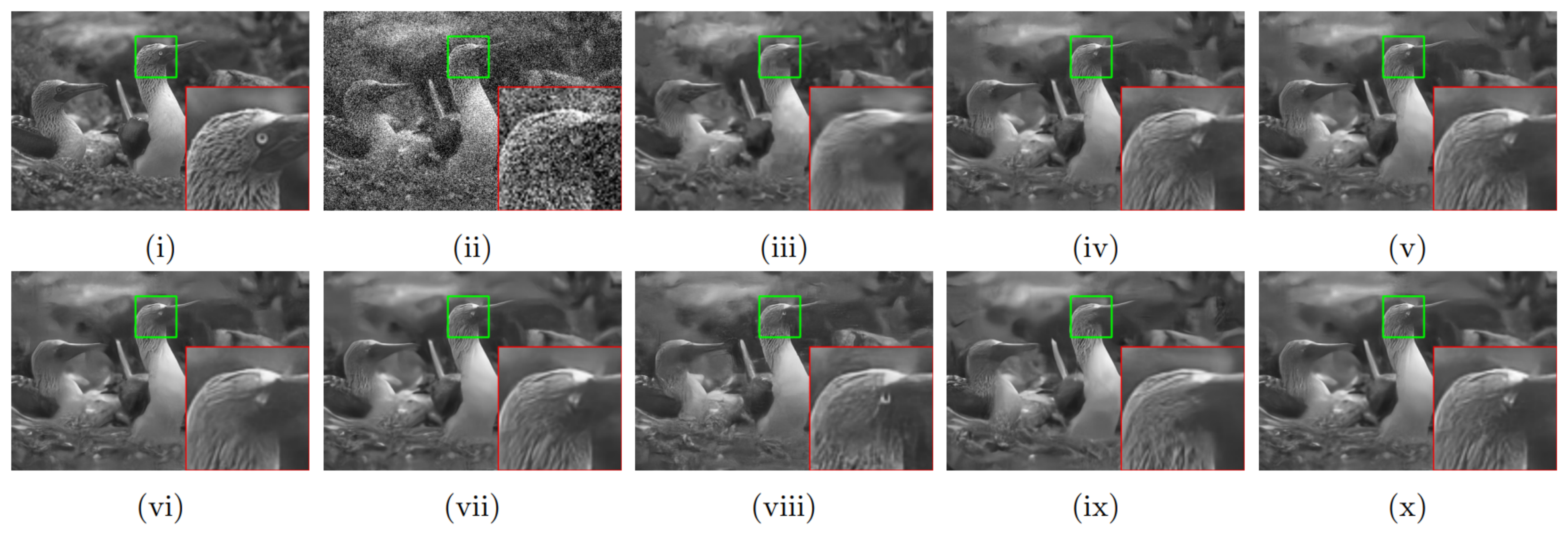}
		\caption{Qualitative comparison between different methods on the image ``test004''. (\romannumeral1) Original / PSNR (dB), (\romannumeral2) Noisy / 14.15, (\romannumeral3) BM3D / 27.27, (\romannumeral4) IRCNN / 27.51, (\romannumeral5) DnCNN-S / 27.60, (\romannumeral6) DnCNN-B / 27.57, (\romannumeral7) FFDNet / 27.70, (\romannumeral8) BUIFD / 27.17, (\romannumeral9) ADNet / 27.64, (\romannumeral10) TSP-RDANet / 27.70.}
        \label{fig:test004}
	\end{center}
\end{figure*}

The visual comparison was also implemented in our experiments. Fig. \ref{fig:test004} shows the comparison between different denoising methods on synthetic noise removal of a grayscale image, where the image used for evaluation is the ``test004'' from the BSD68 dataset. The image was contaminated by the AWGN with a standard deviation of 50. We zoom in a region (green box) for visual detail comparison (red box). One can find that the TSP-RDANet model can remove noise more effectively, meanwhile the model can keep more details of the image. It can be verified that the proposed TSP-RDANet can generate robust denoising performance in grayscale image denoising, both subjectively and objectively.

\subsubsection{Color image denoising evaluation}
For the synthetic noise removal evaluation on color images, we adopted three public and commonly used datasets, including the CBSD68 \cite{Roth2005}, Kodak24 \cite{Kodak24}, and McMaster \cite{Zhang2011} datasets. These datasets were polluted by the AWGN with standard deviations of 15, 25, and 50. Table \ref{tab:CBSD68}, Table \ref{tab:Kodak24}, and Table \ref{tab:McMaster} list the averaged PSNR and SSIM results of our TSP-RDANet and the compared denoising methods, and the red and blue numbers are the best and second-best noise reduction results, respectively. Similar with the performance on grayscale images, the proposed model can obtain competitive results at noise level 15. At the noise level of 25 and 50, our model achieves the highest denoising performance.

\begin{table*}[htbp]
\centering
\caption{The PSNR and SSIM comparisons of multiple denoising models on the CBSD68 dataset.}
\label{tab:CBSD68}
\begin{tabular}{ccccc}
\cline{1-5}
Metrics &  Models & $\sigma$=15 & $\sigma$=25 & $\sigma$=50 \\
\cline{1-5}
\multirow{13}*{PSNR} & CBM3D \cite{Dabov2007} & 33.52 & 30.71 & 27.38\\
\cline{2-5}
    & IRCNN \cite{ZhangZGZ2017} & 33.86 & 31.16 & 27.86\\
\cline{2-5}
    & CDnCNN-S \cite{Zhang2017} & 33.89 & 31.23 & 27.92 \\
\cline{2-5}
    & FFDNet \cite{Zhang2018} & 33.87 & 31.21 & 27.96\\
\cline{2-5}
    & BUIFD \cite{Helou2020} & 33.65 & 30.76 & 26.61 \\
\cline{2-5}
    & DSNetB \cite{Peng2019} & 33.91 & 31.28 & 28.05\\
\cline{2-5}
    & VDN \cite{Yue2019} & 33.90 & 31.35 & \textcolor{blue}{28.19}\\
\cline{2-5}
    & RIDNet \cite{Anwar2019} & 34.01 & 31.37 & 28.14\\
\cline{2-5}
    & ADNet \cite{TianX2020} & 33.99 & 31.31 & 28.04\\
\cline{2-5}
    & BRDNet \cite{Tian2020} & 34.10 & 31.43 & 28.16 \\
\cline{2-5}
    & DudeNet \cite{Tian2021} & 34.01 & 31.34 & 28.09\\
\cline{2-5}
    & AirNet \cite{Li2022} & 33.92 & 31.26 & 28.01\\
\cline{2-5}
    & MWDCNN \cite{Tian2023} & \textcolor{red}{34.18} & \textcolor{blue}{31.45} & 28.13\\
\cline{2-5}
    & TSP-RDANet & \textcolor{blue}{34.14} & \textcolor{red}{31.52} & \textcolor{red}{28.33}\\
\cline{1-5}
\multirow{7}*{SSIM} & IRCNN  \cite{ZhangZGZ2017} & 0.929 & 0.882 & 0.790\\
\cline{2-5}
    & CDnCNN-B \cite{Zhang2017} & 0.929 & 0.883 & 0.790\\
\cline{2-5}
    & FFDNet \cite{Zhang2018} & 0.929 & 0.882 & 0.789 \\
\cline{2-5}
    & ADNet \cite{TianX2020}  & \textcolor{red}{0.933} & \textcolor{red}{0.889} & 0.797\\
\cline{2-5}
    & BUIFD \cite{Helou2020} & 0.930 & 0.882 & 0.777\\
\cline{2-5}
    & AirNet \cite{Li2022} & \textcolor{red}{0.933} & \textcolor{blue}{0.888} & \textcolor{blue}{0.798}\\
\cline{2-5}
    & TSP-RDANet & \textcolor{blue}{0.932} & \textcolor{red}{0.889} & \textcolor{red}{0.804}\\
\cline{1-5}
\end{tabular}
\end{table*}

\begin{table*}[htbp]
\centering
\caption{The PSNR and SSIM comparisons of different models on the Kodak24 dataset.}
\label{tab:Kodak24}
\begin{tabular}{ccccc}
\cline{1-5}
Metrics &  Models & $\sigma$=15 & $\sigma$=25 & $\sigma$=50 \\
\cline{1-5}
\multirow{12}*{PSNR} & CBM3D \cite{Dabov2007} & 34.28 & 31.68 & 28.46\\
\cline{2-5}
    & IRCNN \cite{ZhangZGZ2017} & 34.56 & 32.03 & 28.81\\
\cline{2-5}
    & CDnCNN-S \cite{Zhang2017} & 34.48 & 32.03 & 28.85 \\
\cline{2-5}
    & FFDNet \cite{Zhang2018} & 34.63 & 32.13 & 28.98 \\
\cline{2-5}
& BUIFD \cite{Helou2020} & 34.41 & 31.77 & 27.74\\
\cline{2-5}
    & DSNetB \cite{Peng2019} & 34.63 & 32.16 & 29.05 \\
\cline{2-5}
    & ADNet \cite{TianX2020} & 34.76 & 32.26 & 29.10 \\
\cline{2-5}
    & BRDNet \cite{Tian2020} & 34.88 & \textcolor{blue}{32.41} & 29.22 \\
\cline{2-5}
    & DudeNet \cite{Tian2021} & 34.81 & 32.26 & 29.10 \\
\cline{2-5}
    & AirNet \cite{Li2022} & 34.68 & 32.21 & 29.06\\
\cline{2-5}
    & MWDCNN \cite{Tian2023} & \textcolor{blue}{34.91} & 32.40 & \textcolor{blue}{29.26}\\
\cline{2-5}
    & TSP-RDANet & \textcolor{red}{34.98} & \textcolor{red}{32.54} & \textcolor{red}{29.46} \\
\cline{1-5}
\multirow{7}*{SSIM} & IRCNN  \cite{ZhangZGZ2017} & 0.920 & 0.877 & 0.793\\
\cline{2-5}
    & CDnCNN-B \cite{Zhang2017} & 0.920 & 0.876 & 0.791\\
\cline{2-5}
    & FFDNet \cite{Zhang2018} & 0.922 & 0.878 & 0.794\\
\cline{2-5}
    & ADNet \cite{TianX2020}  & 0.924 & 0.882 & 0.798\\
\cline{2-5}
    & BUIFD \cite{Helou2020} & 0.923 & 0.879 & 0.786\\
\cline{2-5}
    & AirNet \cite{Li2022} & 0.924 & 0.882 & 0.799\\
\cline{2-5}
    & MWDCNN \cite{Tian2023} & \textcolor{red}{0.927} & \textcolor{blue}{0.886} & \textcolor{blue}{0.806}\\
\cline{2-5}
    & TSP-RDANet & \textcolor{blue}{0.926} & \textcolor{red}{0.887} & \textcolor{red}{0.812}\\
\cline{1-5}
\end{tabular}
\end{table*}

\begin{table*}[htbp]
\centering
\caption{The PSNR and SSIM comparisons of image denoising models on the McMaster dataset.}
\label{tab:McMaster}
\begin{tabular}{ccccc}
\cline{1-5}
Metrics &  Models & $\sigma$=15 & $\sigma$=25 & $\sigma$=50 \\
\cline{1-5}
\multirow{11}*{PSNR} & CBM3D \cite{Dabov2007} & 34.06 & 31.66 & 28.51 \\
\cline{2-5}
    & IRCNN \cite{ZhangZGZ2017} & 34.58 & 32.18 & 28.91\\
\cline{2-5}
    & CDnCNN-S \cite{Zhang2017} & 33.44 & 31.51 & 28.61\\
\cline{2-5}
    & FFDNet \cite{Zhang2018} & 34.66 & 32.35 & 29.18\\
\cline{2-5}
& BUIFD \cite{Helou2020} & 33.84 & 31.06 & 26.20\\
\cline{2-5}
    & DSNetB \cite{Peng2019} & 34.67 & 32.40 & 29.28\\
\cline{2-5}
    & ADNet \cite{TianX2020} & 34.93 & 32.56  & 29.36 \\
\cline{2-5}
    & BRDNet \cite{Tian2020} & \textcolor{red}{35.08} & \textcolor{blue}{32.75} & \textcolor{blue}{29.52} \\
\cline{2-5}
    & AirNet \cite{Li2022} & 34.70 & 32.44 & 29.26 \\
\cline{2-5}
    & TSP-RDANet & \textcolor{blue}{35.06} & \textcolor{red}{32.81} & \textcolor{red}{29.74} \\
\cline{1-5}
\multirow{7}*{SSIM} & IRCNN  \cite{ZhangZGZ2017} & 0.920 & 0.882 & 0.807\\
\cline{2-5}
    & CDnCNN-B \cite{Zhang2017} & 0.904 & 0.869 & 0.799\\
\cline{2-5}
    & FFDNet \cite{Zhang2018} & 0.922 & 0.886 & 0.815 \\
\cline{2-5}
    & ADNet \cite{TianX2020}  & \textcolor{blue}{0.927} & \textcolor{blue}{0.894} & \textcolor{blue}{0.825}\\
\cline{2-5}
    & BUIFD \cite{Helou2020} & 0.901 & 0.847 & 0.733\\
\cline{2-5}
    & AirNet \cite{Li2022} & 0.925 & 0.891 & 0.822\\
\cline{2-5}
    & TSP-RDANet & \textcolor{red}{0.928} & \textcolor{red}{0.897} & \textcolor{red}{0.836}\\
\cline{1-5}
\end{tabular}
\end{table*}

\begin{figure*}[htbp]
	\begin{center}
		\includegraphics[width=\textwidth]{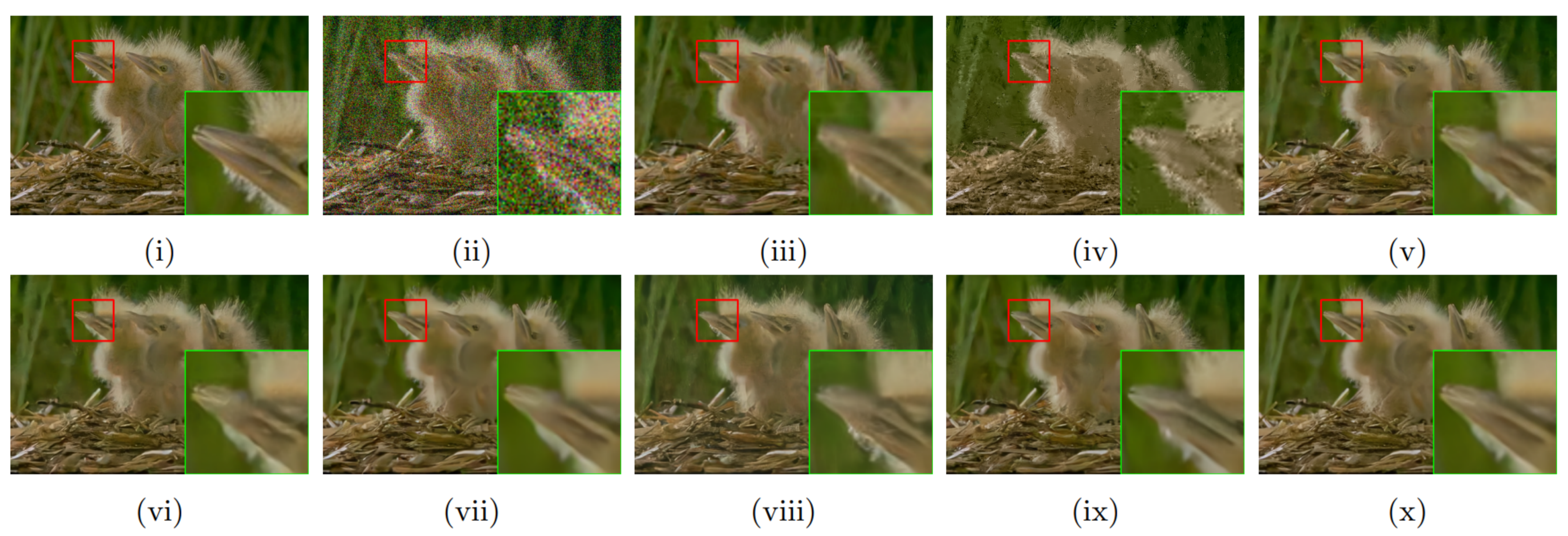}
		\caption{Qualitative comparison on the image ``163085''. (\romannumeral1) Original / PSNR (dB), (\romannumeral2) Noisy / 14.15, (\romannumeral3) CBM3D / 28.34, (\romannumeral4) MCWNNM / 23.85, (\romannumeral5) IRCNN / 28.69, (\romannumeral6) CDnCNN-B / 28.68, (\romannumeral7) FFDNet / 28.75, (\romannumeral8) BUIFD / 26.88, (\romannumeral9) ADNet / 28.80, (\romannumeral10) TSP-RDANet / 29.01.}
        \label{fig:163085}
	\end{center}
\end{figure*}

\begin{figure*}[htbp]
	\begin{center}
		\includegraphics[width=\textwidth]{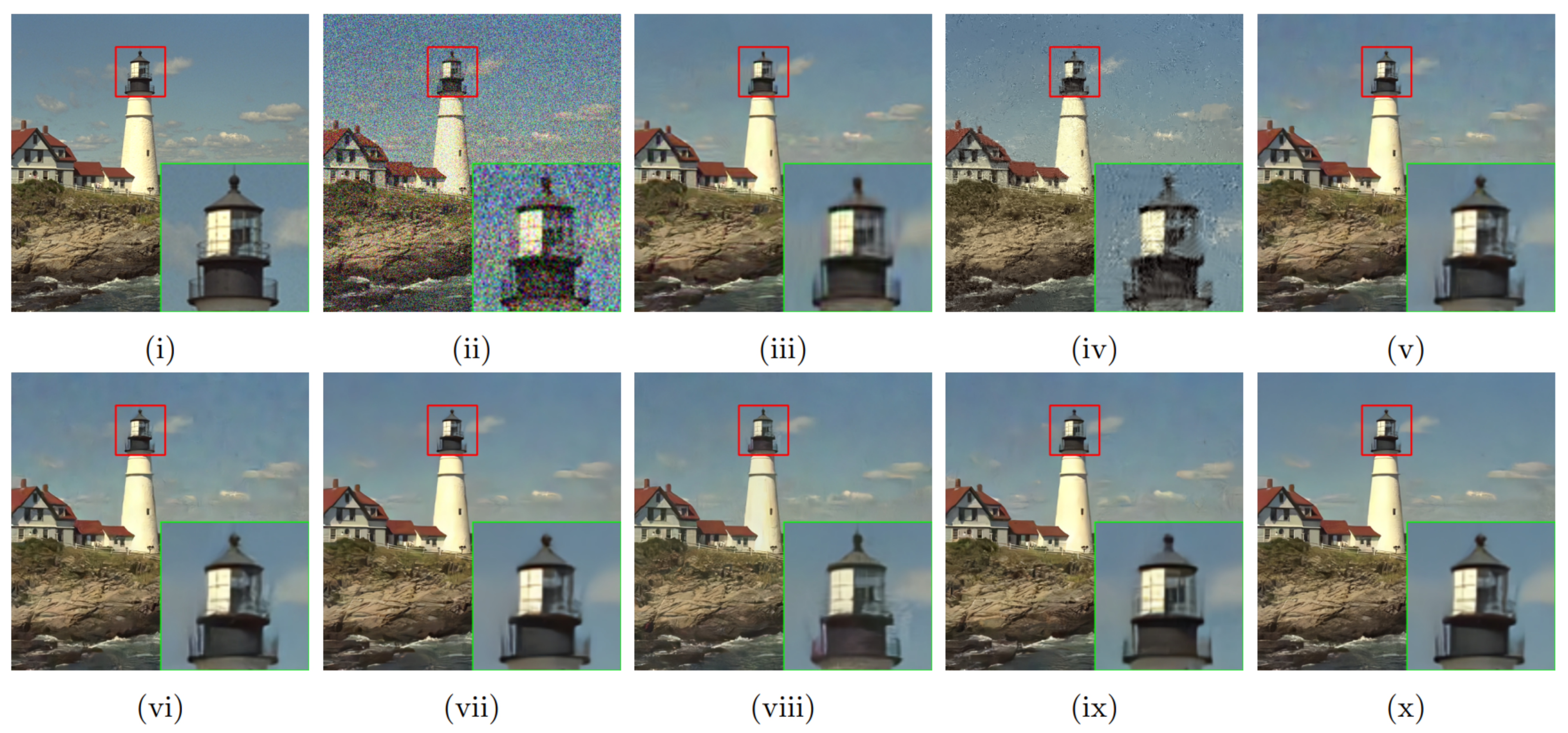}
		\caption{Qualitative comparison on the image ``kodim21''. (\romannumeral1) Original / PSNR (dB), (\romannumeral2) Noisy / 14.15, (\romannumeral3) CBM3D / 27.44, (\romannumeral4) MCWNNM / 23.35, (\romannumeral5) IRCNN / 27.97, (\romannumeral6) CDnCNN-B / 28.03, (\romannumeral7) FFDNet / 28.07, (\romannumeral8) BUIFD / 27.61, (\romannumeral9) ADNet / 28.15, (\romannumeral10) TSP-RDANet / 28.28.}
        \label{fig:kodim21}
	\end{center}
\end{figure*}

The qualitative comparison was also implemented for color image denoising. Fig. \ref{fig:163085} and Fig. \ref{fig:kodim21} display the visual assessment of our TSP-RDANet and other models on synthetic color image denoising. The results on the ``163085'' image from the CBSD68 dataset and ``kodim21'' image from the Kodak24 dataset are presented. These images were contaminated by the AWGN with the noise level of 50. We zoom in an image region (red box) for a more detailed comparison (green box). One can find from Fig. \ref{fig:163085} and Fig. \ref{fig:kodim21} that the TSP-RDANet obtains a desirable trade-off between noise reduction and image texture retaining.

It should be noted from Tables \ref{tab:Set12_PSNR}-\ref{tab:McMaster} that with the increase of noise level, the denoising result of the TSP-RDANet is better than other compared models. This is due to the fact that the designed RDAM and HDRDAM can greatly increase the receptive field of the TSP-RDANet, which helps it extract more contextual information and cope with stronger noise \cite{Zhang2017, Zhang2018}.

\subsection{Real-world noise reduction evaluation}\label{RNRE}
For real-world noise reduction assessment, we utilized the SIDD validation set \cite{Abdelhamed2018} and the DND sRGB images \cite{Plotz2017} as the test datasets. The classical and the state-of-the-art blind denoising models, such as the TWSC \cite{Xu2018}, MCWNNM \cite{Xu2017}, DnCNN-B \cite{Zhang2017}, RIDNet \cite{Anwar2019}, CBDNet \cite{Guo2019}, VDN \cite{Yue2019}, AINDNet \cite{Kim2020}, DANet+ \cite{Yue2020}, DeamNet \cite{Ren2021}, VDIR \cite{Soh2022}, SADNet \cite{Chang2020}, and CycleISP \cite{Zamir2020}, were applied for comparing. Table \ref{tab:SIDD_DND} lists the averaged PSNR and SSIM values of our TSP-RDANet and other denoising methods, and the best and the second performances are emphasized in red and blue, respectively. One can find that the TSP-RDANet obtains the leading results compared with other methods. On the SIDD validation set, the proposed TSP-RDANet achieves 0.30 dB, 0.63 dB, 0.15 dB, 0.23 dB, 0.32 dB, 0.12 dB, and 0.06 dB performance improvement on the state-of-the-art methods VDN, AINDNet, DANet+, DeamNet, VDIR, SADNet, and CycleISP, respectively.

\begin{table*}[htbp]
\centering
\caption{The performance (PSNR/SSIM) comparison between different models on the SIDD and DND datasets.}
\label{tab:SIDD_DND}
\begin{tabular}{ccc}
\hline
Models & SIDD & DND\\
\hline
TWSC \cite{Xu2018} & 35.33/0.933 & 37.94/0.940\\
\hline
MCWNNM \cite{Xu2017} & 33.40/0.879 & 37.38/0.929\\
\hline
DnCNN-B \cite{Zhang2017} & 23.66/0.583 & 37.90/0.943\\
\hline
RIDNet \cite{Anwar2019} & 38.71/0.954 & 39.23/0.953\\
\hline
CBDNet \cite{Guo2019} &  30.78/0.951 & 38.06/0.942\\
\hline
VDN \cite{Yue2019} & 39.28/0.956 & 39.38/0.952\\
\hline
AINDNet \cite{Kim2020} & 38.95/0.952 & 39.37/0.951\\
\hline
DANet+ \cite{Yue2020} & 39.43/0.956 & 39.58/\textcolor{blue}{0.955}\\
\hline
DeamNet \cite{Ren2021} & 39.35/0.955 & \textcolor{blue}{39.63}/0.953\\
\hline
VDIR \cite{Soh2022} & 39.26/0.955 & \textcolor{blue}{39.63}/0.953\\
\hline
SADNet \cite{Chang2020} & 39.46/\textcolor{blue}{0.957} & 39.59/0.952\\
\hline
CycleISP \cite{Zamir2020} & \textcolor{blue}{39.52}/\textcolor{blue}{0.957} & 39.56/\textcolor{red}{0.956}\\
\hline
TSP-RDANet & \textcolor{red}{39.58}/\textcolor{red}{0.958} & \textcolor{red}{39.70}/0.954\\
\hline
\end{tabular}
\end{table*}

\begin{figure*}[htbp]
	\begin{center}
		\includegraphics[width=\textwidth]{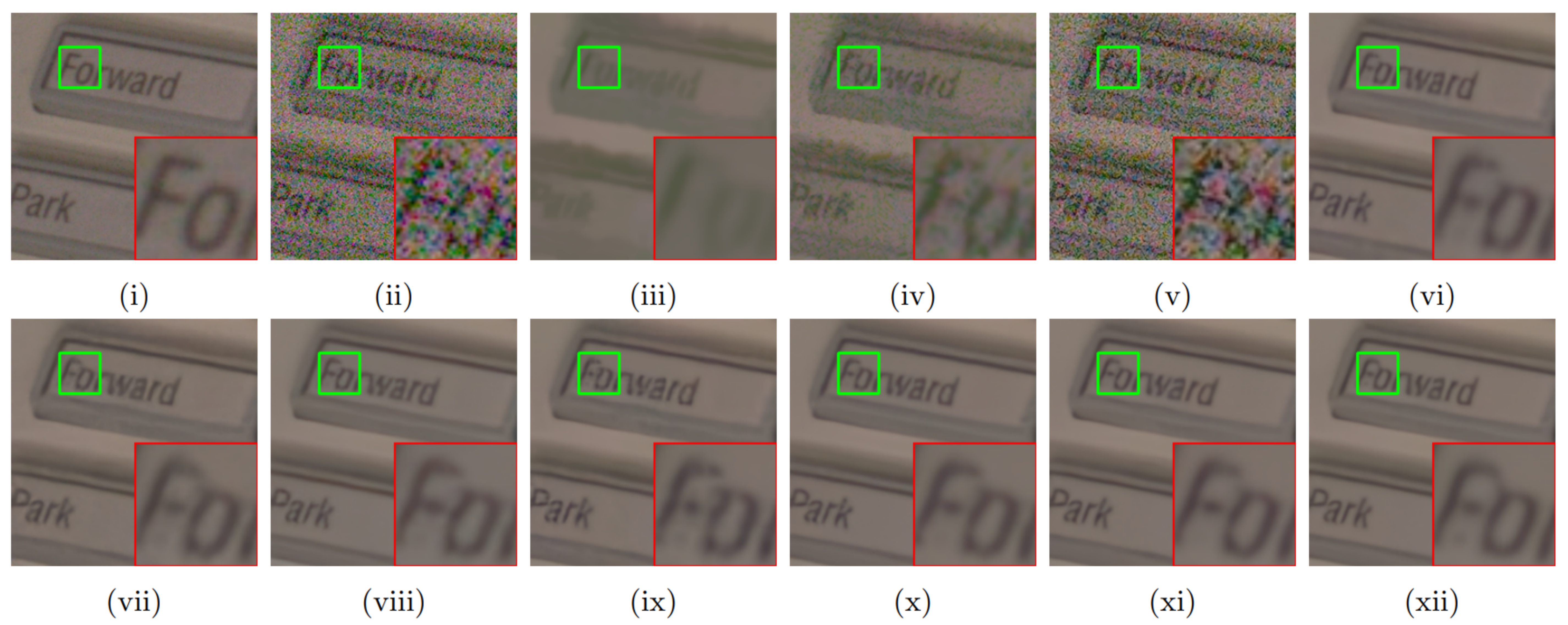}
		\caption{Qualitative comparison of different denoising models on the image ``10\_3''. (\romannumeral1) Original / PSNR (dB), (\romannumeral2) Noisy / 18.25, (\romannumeral3) TWSC / 30.42, (\romannumeral4) MCWNNM / 28.63, (\romannumeral5) CDnCNN-B / 20.76, (\romannumeral6) VDN / 36.39, (\romannumeral7) AINDNet / 36.24, (\romannumeral8) DANet+ / 36.37, (\romannumeral9) VDIR / 36.35, (\romannumeral10) SADNet / 36.72, (\romannumeral11) CycleISP / 36.72, (\romannumeral12) TSP-RDANet / 36.84.}
        \label{fig:10_3}
	\end{center}
\end{figure*}

Fig. \ref{fig:10_3} shows the visual denoising results of the proposed TSP-RDANet and other compared models for real image denoising. The image ``10\_3'' from the SIDD was used for visual presentation. An image region (green box) is enlarged for a more detailed visual comparison (red box). As can be seen from Fig. \ref{fig:10_3}, the TSP-RDANet obtains a more visually appealing result. Moreover, the TSP-RDANet also surpasses all other denoising models on PSNR value.

\subsection{Model computational complexity evaluation}
The model complexity of all the compared models was also evaluated in our experiments, which was quantified by the running time and network parameters. The BM3D, TWSC, and MCWNNM were performed in the Matlab (R2020a) platform, and the remaining methods were executed in the PyCharm (2021) software. We randomly selected three clean color images with the sizes of $256\times256$, $512\times512$, and $1024\times1024$ to obtain the runtime of every model on each image. We performed single-channel processing of these color images to obtain their grayscale counterparts. The clean grayscale and color images were then added to the AWGN at noise level 25 to generate the noisy images for evaluation. The public pytorch-OpCounter package\footnote{\url{https://github.com/Lyken17/pytorch-OpCounter}} was used to obtain the model parameters of the evaluated denoising models. In order to make a more precise and objective comparison, we ran 20 executions of each denoising model to obtain the average runtime, which was used for comparing the complexity of different denoising models.

Table \ref{tab:time} reports the runtime of all tested models. It can be seen that the running speeds of our TSP-RDANet outperform the BM3D, TWSC, MCWNNM, AINDNet, and VDIR at all image sizes. Although the denoising speed of the proposed TSP-RDANet is slower than the IRCNN, DnCNN-B, FFDNet, BUIFD, DANet+, CycleISP, ADNet, and SADNet, the denoising effect of our model outperforms these models. The numbers of the network parameters of different models were offered in Table \ref{tab:Parameters}. One can find that the proposed TSP-RDANet owns fewer network parameters than many denoising methods such as the DANet+, VDN, SADNet, AINDNet, CBDNet, and AirNet. Meanwhile, our model can achieve better quantitative and qualitative denoising results than these models (see Sec. \ref{SNRE} and Sec. \ref{RNRE}). It can be verified from our experiments that our model obtains a favorable balance between denoising result and complexity.

\begin{table*}[htbp]
\centering
\caption{Runtime results (in seconds) of the compared models on different synthetic grayscale and color noisy images.}
\label{tab:time}
\begin{tabular}{cccccccc}
\cline{1-8}
\multirow{2}*{Devices} & \multirow{2}*{Models}	& \multicolumn{2}{c}{$256\times256$}	& \multicolumn{2}{c}{$512\times512$}	& \multicolumn{2}{c}{$1024\times1024$}\\
\cline{3-8}
 \multicolumn{1}{c}{} &  & Grayscale & Color & Grayscale & Color & Grayscale & Color\\
\cline{1-8}
\multirow{2}*{CPU} & BM3D \cite{Dabov2007} & 0.458	& 0.593	& 2.354	& 3.771	& 9.782	& 12.818\\
\cline{2-8}
    & TWSC \cite{Xu2018} & 12.314 & 34.41 & 53.155 & 140.964 & 221.507 & 608.492 \\
\cline{2-8}
    & MCWNNM \cite{Xu2017} & - & 62.777 & - & 277.623 & - & 1120.112  \\
\cline{1-8}
\multirow{14}*{GPU} & IRCNN \cite{ZhangZGZ2017}  & 0.030	& 0.030	& 0.030	& 0.030	& 0.030	& 0.030\\
\cline{2-8}
    & DnCNN-B \cite{Zhang2017} & 0.032	& 0.032	& 0.037	& 0.037	& 0.057	& 0.057\\
\cline{2-8}
    & FFDNet \cite{Zhang2018} & 0.031	& 0.030	& 0.031	& 0.030	& 0.032	& 0.030\\
\cline{2-8}
    & BUIFD \cite{Helou2020} & 0.035	& 0.037	& 0.050	& 0.053	& 0.112	& 0.123\\
\cline{2-8}
    & DANet+ \cite{Yue2020}  &  - & 0.025 & - & 0.027 & - & 0.041\\
\cline{2-8}
    & ADNet \cite{TianX2020} & 0.031 & 0.033 & 0.035 & 0.045 & 0.051 & 0.093\\
\cline{2-8}
    & SADNet \cite{Chang2020}  & 0.030 & 0.030 & 0.043 & 0.044 & 0.101 & 0.102\\
\cline{2-8}
    & CycleISP \cite{Zamir2020}  &  - & 0.055 & - & 0.156 & - & 0.545\\
\cline{2-8}
    & AirNet \cite{Li2022}  &  - & 0.143 & - & 0.498 & - & 2.501\\
\cline{2-8}
    & VDN \cite{Yue2019} & 0.144 & 0.162 & 0.607 & 0.597 & 2.367 & 2.376\\
\cline{2-8}
    & AINDNet \cite{Kim2020} & - & 0.531 & - & 2.329 & - & 9.573\\
\cline{2-8}
    & VDIR\cite{Soh2022}  &  - & 0.385 & - & 1.622 & - & 6.690\\
\cline{2-8}
    & TSP-RDANet & 0.261 & 0.263 & 0.625 & 0.627 & 1.979 & 2.004\\
\cline{1-8}
\end{tabular}
\end{table*}

\begin{table*}[htbp]
\centering
\caption{Number of parameters (in K) of the compared image denoising models.}
\label{tab:Parameters}
\begin{tabular}{ccc}
\hline
\multirow{2}*{Models} & \multicolumn{2}{c}{Number of model parameters} \\
\cline{2-3}
 \multicolumn{1}{c}{}  & Grayscale & Color \\
\hline
IRCNN \cite{ZhangZGZ2017} & 186 & 188 \\
\hline
DnCNN-B \cite{Zhang2017} & 668 & 673 \\
\hline
FFDNet \cite{Zhang2018} & 485 & 852 \\
\hline
BUIFD \cite{Helou2020} & 1075 & 1085 \\
\hline
RIDNet \cite{Anwar2019} & 1497 & 1499 \\
\hline
BRDNet \cite{Tian2020} & 1113 & 1117 \\
\hline
ADNet \cite{TianX2020} & 519 & 521 \\
\hline
DudeNet \cite{Tian2021} & 1077 & 1079 \\
\hline
DeamNet \cite{Ren2021} & 1873 & 1876 \\
\hline
VDIR \cite{Soh2022}  & - & 2227 \\
\hline
CycleISP \cite{Zamir2020} & - & 2837 \\
\hline
CBDNet \cite{Guo2019} & - & 4365 \\
\hline
VDN \cite{Yue2019} & 7810 & 7817 \\
\hline
DANet+ \cite{Yue2020} & - & 9154 \\
\hline
AINDNet \cite{Kim2020} & - & 13764 \\
\hline
SADNet \cite{Chang2020} & 3450 & 3451 \\
\hline
AirNet \cite{Li2022} & - & 8930 \\
\hline
TSP-RDANet & 2846 & 2848 \\
\hline
\end{tabular}
\end{table*}

\section{Conclusion}\label{Conclusion}
In this work, we introduce a new two-stage progressive network (TSP-RDANet) for image noise removal. The proposed model decomposes the entire denoising process into two sub-stage tasks to progressively obtain more effective noise reduction. Especially, we design the novel residual dense attention module (RDAM) for the first stage of the model. The hybrid dilated residual dense attention module (HDRDAM) is developed for the second stage. The RDAM and HDRDAM consist of the dense block, attention block, and residual learning, which makes them be capable of extracting rich local features and filtering irrelevant and redundant features to further improve their learning ability. The long skip connection is utilized to link the two sub-networks to achieve progressive image denoising. Downsampling and the dilated convolution are applied to the TSP-RDANet model to enlarge its receptive field size and further capture more contextual information to enhance its restoration capability. Compared with the state-of-the-art models, our TSP-RDANet achieves excellent and competitive denoising results on multiple denoising tasks.

The experimental results yet also reveal that the discriminative ability of the proposed model under weak noise levels still needs to be further investigated. Moreover, although the proposed model can produce competitive blind denoising performance, the model needs clean/noisy image pairs to train. In our following work, we will focus on evolving the model into a self-supervised or unsupervised manner.

\section{Acknowledgements}
This work is supported by the Natural Science Foundation of China (61863037, 41971392) and the Applied Basic Research Foundation of Yunnan Province (202001AT070077).

\bibliography{reference}

\begin{thebibliography}{10}
\expandafter\ifx\csname url\endcsname\relax
  \def\url#1{\texttt{#1}}\fi
\expandafter\ifx\csname urlprefix\endcsname\relax\def\urlprefix{URL }\fi
\expandafter\ifx\csname href\endcsname\relax
  \def\href#1#2{#2} \def\path#1{#1}\fi

\bibitem{Dabov2007}
K.~Dabov, A.~Foi, V.~Katkovnik, K.~O. Egiazarian, Image denoising by sparse 3-d
  transform-domain collaborative filtering, {IEEE} Trans. Image Process. 16~(8)
  (2007) 2080--2095.

\bibitem{Xu2018}
J.~Xu, L.~Zhang, D.~Zhang, A trilateral weighted sparse coding scheme for
  real-world image denoising, European Conference on Computer Vision 11212
  (2018) 21--38.

\bibitem{Ou2022}
Y.~Ou, M.~N.~S. Swamy, J.~Luo, B.~Li, Single image denoising via multi-scale
  weighted group sparse coding, Signal Process. 200 (2022) 108650.

\bibitem{Xu2017}
J.~Xu, L.~Zhang, D.~Zhang, X.~Feng, Multi-channel weighted nuclear norm
  minimization for real color image denoising, {IEEE} International Conference
  on Computer Vision (2017) 1105--1113.

\bibitem{Zhang2017}
K.~Zhang, W.~Zuo, Y.~Chen, D.~Meng, L.~Zhang, Beyond a gaussian denoiser:
  Residual learning of deep {CNN} for image denoising, {IEEE} Trans. Image
  Process. 26~(7) (2017) 3142--3155.

\bibitem{Peng2019}
Y.~Peng, L.~Zhang, S.~Liu, X.~Wu, Y.~Zhang, X.~Wang, Dilated residual networks
  with symmetric skip connection for image denoising, Neurocomputing 345 (2019)
  67--76.

\bibitem{Zhang2021}
Y.~Zhang, Y.~Tian, Y.~Kong, B.~Zhong, Y.~Fu, Residual dense network for image
  restoration, {IEEE} Trans. Pattern Anal. Mach. Intell. 43~(7) (2021)
  2480--2495.

\bibitem{Jia2021}
F.~Jia, W.~H. Wong, T.~Zeng, Ddunet: Dense dense u-net with applications in
  image denoising, in: International Conference on Computer Vision Workshops,
  2021, pp. 354--364.

\bibitem{TianX2020}
C.~Tian, Y.~Xu, Z.~Li, W.~Zuo, L.~Fei, H.~Liu, Attention-guided {CNN} for image
  denoising, Neural Networks 124 (2020) 117--129.

\bibitem{Anwar2019}
S.~Anwar, N.~Barnes, Real image denoising with feature attention, International
  Conference on Computer Vision (2019) 3155--3164.

\bibitem{Zamir2020}
S.~W. Zamir, A.~Arora, S.~H. Khan, M.~Hayat, F.~S. Khan, M.~Yang, L.~Shao,
  Cycleisp: Real image restoration via improved data synthesis, in: {IEEE}
  Conference on Computer Vision and Pattern Recognition, 2020, pp. 2693--2702.

\bibitem{Bai2023}
Y.~Bai, M.~Liu, C.~Yao, C.~Lin, Y.~Zhao, Mspnet: Multi-stage progressive
  network for image denoising, Neurocomputing 517 (2023) 71--80.

\bibitem{Wang2023}
Z.~Wang, Y.~Bai, Y.~Zhou, C.~Xie, Can cnns be more robust than transformers?,
  in: International Conference on Learning Representations, 2023.

\bibitem{Krizhevsky2012}
A.~Krizhevsky, I.~Sutskever, G.~E. Hinton, Imagenet classification with deep
  convolutional neural networks, Advances in Neural Information Processing
  Systems (2012) 1106--1114.

\bibitem{Ioffe2015}
S.~Ioffe, C.~Szegedy, Batch normalization: Accelerating deep network training
  by reducing internal covariate shift, International Conference on Machine
  Learning 37 (2015) 448--456.

\bibitem{He2016}
K.~He, X.~Zhang, S.~Ren, J.~Sun, Deep residual learning for image recognition,
  {IEEE} Conference on Computer Vision and Pattern Recognition (2016) 770--778.

\bibitem{ZhangZGZ2017}
K.~Zhang, W.~Zuo, S.~Gu, L.~Zhang, Learning deep {CNN} denoiser prior for image
  restoration, {IEEE} Conference on Computer Vision and Pattern Recognition
  (2017) 2808--2817.

\bibitem{Yu2015}
F.~Yu, V.~Koltun, Multi-scale context aggregation by dilated convolutions, in:
  International Conference on Learning Representations, 2016.

\bibitem{Zhang2018}
K.~Zhang, W.~Zuo, L.~Zhang, Ffdnet: Toward a fast and flexible solution for
  cnn-based image denoising, {IEEE} Trans. Image Process. 27~(9) (2018)
  4608--4622.

\bibitem{Yue2019}
Z.~Yue, H.~Yong, Q.~Zhao, D.~Meng, L.~Zhang, Variational denoising network:
  Toward blind noise modeling and removal, Advances in Neural Information
  Processing Systems (2019) 1688--1699.

\bibitem{Kim2020}
Y.~Kim, J.~W. Soh, G.~Y. Park, N.~I. Cho, Transfer learning from synthetic to
  real-noise denoising with adaptive instance normalization, IEEE Conference on
  Computer Vision and Pattern Recognition (2020) 3479--3489.

\bibitem{Helou2020}
M.~E. Helou, S.~S{\"{u}}sstrunk, Blind universal bayesian image denoising with
  gaussian noise level learning, {IEEE} Trans. Image Process. 29 (2020)
  4885--4897.

\bibitem{Guo2019}
S.~Guo, Z.~Yan, K.~Zhang, W.~Zuo, L.~Zhang, Toward convolutional blind
  denoising of real photographs, {IEEE} Conference on Computer Vision and
  Pattern Recognition (2019) 1712--1722.

\bibitem{Soh2022}
J.~W. Soh, N.~I. Cho, Variational deep image restoration, {IEEE} Trans. Image
  Process. 31 (2022) 4363--4376.

\bibitem{WuS2023}
W.~Wu, S.~Liao, G.~Lv, P.~Liang, Y.~Zhang, Image blind denoising using dual
  convolutional neural network with skip connection (2023).
\newblock \href {http://arxiv.org/abs/2304.01620} {\path{arXiv:2304.01620}}.

\bibitem{Chang2020}
M.~Chang, Q.~Li, H.~Feng, Z.~Xu, Spatial-adaptive network for single image
  denoising, in: European Conference on Computer Vision, Vol. 12375, 2020, pp.
  171--187.

\bibitem{Dai2017}
J.~Dai, H.~Qi, Y.~Xiong, Y.~Li, G.~Zhang, H.~Hu, Y.~Wei, Deformable
  convolutional networks, in: {IEEE} International Conference on Computer
  Vision, 2017, pp. 764--773.

\bibitem{Zhu2019}
X.~Zhu, H.~Hu, S.~Lin, J.~Dai, Deformable convnets {V2:} more deformable,
  better results, in: {IEEE} Conference on Computer Vision and Pattern
  Recognition, 2019, pp. 9308--9316.

\bibitem{Quan2021}
Y.~Quan, Y.~Chen, Y.~Shao, H.~Teng, Y.~Xu, H.~Ji, Image denoising using
  complex-valued deep {CNN}, Pattern Recognit. 111 (2021) 107639.

\bibitem{Li2022}
B.~Li, X.~Liu, P.~Hu, Z.~Wu, J.~Lv, X.~Peng, All-in-one image restoration for
  unknown corruption, {IEEE} Conference on Computer Vision and Pattern
  Recognition (2022) 17431--17441.

\bibitem{Pan2022}
J.~Pan, D.~Sun, J.~Zhang, J.~Tang, J.~Yang, Y.~Tai, M.~Yang, Dual convolutional
  neural networks for low-level vision, Int. J. Comput. Vis. 130~(6) (2022)
  1440--1458.

\bibitem{Ioffe2017}
S.~Ioffe, Batch renormalization: Towards reducing minibatch dependence in
  batch-normalized models, in: Advances in Neural Information Processing
  Systems, 2017, pp. 1945--1953.

\bibitem{Tian2020}
C.~Tian, Y.~Xu, W.~Zuo, Image denoising using deep {CNN} with batch
  renormalization, Neural Networks 121 (2020) 461--473.

\bibitem{Tian2021}
C.~Tian, Y.~Xu, W.~Zuo, B.~Du, C.~Lin, D.~Zhang, Designing and training of a
  dual {CNN} for image denoising, Knowl. Based Syst. 226 (2021) 106949.

\bibitem{Yue2020}
Z.~Yue, Q.~Zhao, L.~Zhang, D.~Meng, Dual adversarial network: Toward real-world
  noise removal and noise generation, in: European Conference on Computer
  Vision, Vol. 12355, 2020, pp. 41--58.

\bibitem{Jiang2023}
B.~Jiang, J.~Li, H.~Li, R.~Li, D.~Zhang, G.~Lu, Enhanced frequency fusion
  network with dynamic hash attention for image denoising, Inf. Fusion 92
  (2023) 420--434.

\bibitem{Ren2021}
C.~Ren, X.~He, C.~Wang, Z.~Zhao, Adaptive consistency prior based deep network
  for image denoising, {IEEE} Conference on Computer Vision and Pattern
  Recognition (2021) 8596--8606.

\bibitem{Woo2018}
S.~Woo, J.~Park, J.~Lee, I.~S. Kweon, {CBAM:} convolutional block attention
  module, European Conference on Computer Vision 11211 (2018) 3--19.

\bibitem{Hu2018}
J.~Hu, L.~Shen, G.~Sun, Squeeze-and-excitation networks, {IEEE} Conference on
  Computer Vision and Pattern Recognition (2018) 7132--7141.

\bibitem{WuLZ2023}
W.~Wu, S.~Liu, Y.~Zhou, Y.~Zhang, Y.~Xiang, Dual residual attention network for
  image denoising (2023).
\newblock \href {http://arxiv.org/abs/2305.04269} {\path{arXiv:2305.04269}}.

\bibitem{Huang2022}
J.~Huang, Z.~Zhao, C.~Ren, Q.~Teng, X.~He, A prior-guided deep network for real
  image denoising and its applications, Knowl. Based Syst. 255 (2022) 109776.

\bibitem{Zhuge2023}
R.~Zhuge, J.~Wang, Z.~Xu, Y.~Xu, Single image denoising with a feature-enhanced
  network, Neural Networks 168 (2023) 313--325.

\bibitem{Thakur2023}
R.~K. Thakur, S.~K. Maji, Multi scale pixel attention and feature extraction
  based neural network for image denoising, Pattern Recognit. 141 (2023)
  109603.

\bibitem{Tian2023}
C.~Tian, M.~Zheng, W.~Zuo, B.~Zhang, Y.~Zhang, D.~Zhang, Multi-stage image
  denoising with the wavelet transform, Pattern Recognit. 134 (2023) 109050.

\bibitem{Zheng2019}
Y.~Zheng, X.~Yu, M.~Liu, S.~Zhang, Residual multiscale based single image
  deraining, in: British Machine Vision Conference, 2019, p. 147.

\bibitem{Ren2019}
D.~Ren, W.~Zuo, Q.~Hu, P.~Zhu, D.~Meng, Progressive image deraining networks:
  {A} better and simpler baseline, in: {IEEE} Conference on Computer Vision and
  Pattern Recognition, 2019, pp. 3937--3946.

\bibitem{Nah2017}
S.~Nah, T.~H. Kim, K.~M. Lee, Deep multi-scale convolutional neural network for
  dynamic scene deblurring, in: {IEEE} Conference on Computer Vision and
  Pattern Recognition, 2017, pp. 257--265.

\bibitem{ZhangD2019}
H.~Zhang, Y.~Dai, H.~Li, P.~Koniusz, Deep stacked hierarchical multi-patch
  network for image deblurring, in: {IEEE} Conference on Computer Vision and
  Pattern Recognition, 2019, pp. 5978--5986.

\bibitem{Suin2020}
M.~Suin, K.~Purohit, A.~N. Rajagopalan, Spatially-attentive patch-hierarchical
  network for adaptive motion deblurring, in: {IEEE} Conference on Computer
  Vision and Pattern Recognition, 2020, pp. 3603--3612.

\bibitem{Fu2020}
X.~Fu, B.~Liang, Y.~Huang, X.~Ding, J.~W. Paisley, Lightweight pyramid networks
  for image deraining, {IEEE} Trans. Neural Networks Learn. Syst. 31~(6) (2020)
  1794--1807.

\bibitem{Han1995}
J.~Han, C.~Moraga, The influence of the sigmoid function parameters on the
  speed of backpropagation learning, International Workshop on Artificial
  Neural Networks 930 (1995) 195--201.

\bibitem{Wang2018}
P.~Wang, P.~Chen, Y.~Yuan, D.~Liu, Z.~Huang, X.~Hou, G.~W. Cottrell,
  Understanding convolution for semantic segmentation, {IEEE} Winter Conference
  on Applications of Computer Vision (2018) 1451--1460.

\bibitem{Zamir2021}
S.~W. Zamir, A.~Arora, S.~H. Khan, M.~Hayat, F.~S. Khan, M.~Yang, L.~Shao,
  Multi-stage progressive image restoration, in: {IEEE} Conference on Computer
  Vision and Pattern Recognition, 2021, pp. 14821--14831.

\bibitem{Lai2017}
W.~Lai, J.~Huang, N.~Ahuja, M.~Yang, Deep laplacian pyramid networks for fast
  and accurate super-resolution, {IEEE} Conference on Computer Vision and
  Pattern Recognition (2017) 5835--5843.

\bibitem{Jiang2020}
K.~Jiang, Z.~Wang, P.~Yi, C.~Chen, B.~Huang, Y.~Luo, J.~Ma, J.~Jiang,
  Multi-scale progressive fusion network for single image deraining, {IEEE}
  Conference on Computer Vision and Pattern Recognition (2020) 8343--8352.

\bibitem{Kamgar1999}
B.~Kamgar{-}Parsi, A.~Rosenfeld, Optimally isotropic laplacian operator, {IEEE}
  Trans. Image Process. 8~(10) (1999) 1467--1472.

\bibitem{Wang2004}
Z.~Wang, A.~C. Bovik, H.~R. Sheikh, E.~P. Simoncelli, Image quality assessment:
  from error visibility to structural similarity, {IEEE} Trans. Image Process.
  13~(4) (2004) 600--612.

\bibitem{Lim2017}
B.~Lim, S.~Son, H.~Kim, S.~Nah, K.~M. Lee, Enhanced deep residual networks for
  single image super-resolution, in: {IEEE} Conference on Computer Vision and
  Pattern Recognition Workshops, 2017, pp. 1132--1140.

\bibitem{Roth2005}
S.~Roth, M.~J. Black, Fields of experts: {A} framework for learning image
  priors, {IEEE} Computer Society Conference on Computer Vision and Pattern
  (2005) 860--867.

\bibitem{Kodak24}
R.~Franzen, Kodak lossless true color image suite,
  \url{http://r0k.us/graphics/kodak/} (1999).

\bibitem{Zhang2011}
L.~Zhang, X.~Wu, A.~Buades, X.~Li, {Color demosaicking by local directional
  interpolation and nonlocal adaptive thresholding}, Journal of Electronic
  Imaging 20~(2) (2011) 1--17.

\bibitem{Abdelhamed2018}
A.~Abdelhamed, S.~Lin, M.~S. Brown, A high-quality denoising dataset for
  smartphone cameras, {IEEE} Conference on Computer Vision and Pattern
  Recognition (2018) 1692--1700.

\bibitem{Plotz2017}
T.~Plotz, S.~Roth, Benchmarking denoising algorithms with real photographs, in:
  IEEE Conference on Computer Vision and Pattern Recognition, 2017, pp.
  1586--1595.

\bibitem{Kingma2014}
D.~P. Kingma, J.~Ba, Adam: {A} method for stochastic optimization, in:
  International Conference on Learning Representations, 2015.

\bibitem{Loshchilov2017}
I.~Loshchilov, F.~Hutter, {SGDR:} stochastic gradient descent with warm
  restarts, in: International Conference on Learning Representations, 2017.

\end{thebibliography}
\end{document}